\documentclass[twoside,11pt]{article}

\usepackage{dmlr2e}

\usepackage{booktabs}
\usepackage{multirow}
\usepackage{xcolor}
\usepackage{colortbl}
\usepackage{array}
\usepackage{subcaption}
\usepackage{enumitem}
\usepackage{tikz}
\usetikzlibrary{arrows.meta,positioning,shapes.geometric}

\dmlrheading{1}{2026}{1-\pageref{LastPage}}{2/26}{TBD}{TBD-0000}{Chun et al.}
\def\openreview{\url{https://openreview.net/forum?id=TBD}}

\ShortHeadings{CEI Benchmark: Pragmatic Reasoning Dataset}{Chun et al.}
\firstpageno{1}

\begin{document}

\title{CEI: A Benchmark for Evaluating Pragmatic Reasoning in Language Models}

\author{\name Jon Chun\thanks{Also affiliated with US NIST AI Consortium (PI) and Schmidt Science HAVI (PI).} \email chunj@kenyon.edu \\
       \name Hannah Sussman \email lattermann1@kenyon.edu \\
       \name Adrian Mangine \email mangine1@kenyon.edu \\
       \name Murathan Kocaman \email kocaman1@kenyon.edu \\
       \name Kirill Sidorko \email sidorko1@kenyon.edu \\
       \name Abhigya Koirala \email koirala1@kenyon.edu \\
       \name Andre McCloud \email mccloud1@kenyon.edu \\
       \name Gwen Eisenbeis \email eisenbeis1@kenyon.edu \\
       \name Wisdom Akanwe \email akanwe1@kenyon.edu \\
       \name Moustapha Gassama \email gassama1@kenyon.edu \\
       \name Eliezer Gonzalez Chirinos \email gonzalez4@kenyon.edu \\
       \name Anne-Duncan Enright \email enright1@kenyon.edu \\
       \name Peter Dunson \email dunson1@kenyon.edu \\
       \name Tiffanie Ng \email ng2@kenyon.edu \\
       \name Anna von Rosenstiel \email vonrosenstiel1@kenyon.edu \\
       \name Godwin Idowu \email idowu1@kenyon.edu \\
       \addr Kenyon College, Gambier, OH, USA}

\editor{TBD}

\maketitle

\begin{abstract}
Pragmatic reasoning---inferring intended meaning beyond literal semantics---underpins everyday communication yet remains difficult for large language models. We present the Contextual Emotional Inference (CEI) Benchmark: 300 human-validated scenarios for evaluating how well LLMs disambiguate pragmatically complex utterances. Each scenario pairs a situational context and speaker-listener roles (with explicit power relations) against an ambiguous utterance. The dataset covers five pragmatic subtypes (sarcasm/irony, mixed signals, strategic politeness, passive aggression, deflection/misdirection) drawn from workplace, family, social, and service settings, with three power configurations (peer, higher$\rightarrow$lower, lower$\rightarrow$higher). Three trained annotators independently labeled every scenario. Inter-annotator agreement (Fleiss' $\kappa = 0.06$--$0.25$ by subtype) is low but expected: pragmatic inference admits multiple valid readings, and the disagreement itself is informative. We describe our annotation methodology, including a 4-level quality control pipeline that combines automated statistical checks with expert adjudication. CEI is released under CC-BY-4.0.
\end{abstract}

\begin{keywords}
pragmatic reasoning, benchmark, language models, annotation, social reasoning, emotion inference
\end{keywords}

\section{Introduction}
\label{sec:intro}

When a junior employee tells their manager ``Sure, I'll handle the extra work this weekend,'' the intended meaning depends on context: Is this sincere agreement? Strategic politeness masking reluctance? Passive-aggressive compliance? Recovering the intended meaning from what is literally said (pragmatic reasoning) has been studied extensively in linguistics \citep{grice1975logic,sperber1986relevance}, but remains hard to operationalize as an NLP task.

Language models are now deployed for communication analysis in HR screening, sentiment analysis, and content moderation, where pragmatic interpretation is required \citep{liang2022holistic}. These applications demand that a model integrate contextual cues, social roles, and power dynamics rather than parse surface meaning alone. Existing benchmarks address narrow pragmatic phenomena such as sarcasm detection \citep{ghosh2020semeval} or test adjacent capabilities \citep{sap2019socialiqa,le2019revisiting}, but to the best of our knowledge, none targets single-turn pragmatic reasoning across multiple social contexts and communication styles simultaneously.

We introduce the \textbf{Contextual Emotional Inference (CEI) Benchmark}, a dataset of 300 scenarios spanning 5 pragmatic subtypes and 3 power relations, each labeled independently by 3 trained annotators (900 annotations total). Inter-annotator agreement is low (Fleiss' $\kappa = 0.06$--$0.25$ by subtype), which we argue reflects the difficulty of the task rather than noisy annotation. To catch actual quality problems, we developed a 4-level QA pipeline combining automated statistical checks, agreement analysis, and expert adjudication. We release the dataset, annotation guidelines, and all code under permissive licenses (CC-BY-4.0 for data, MIT for code).\footnote{Code: \url{https://github.com/jon-chun/cei-tom-dataset-base}; Dataset: \url{https://huggingface.co/datasets/jonc/cei-benchmark}}

Our contributions are:
\begin{enumerate}[leftmargin=*,noitemsep]
    \item \textbf{A benchmark dataset} of 300 expert-authored scenarios covering five pragmatic subtypes (sarcasm, mixed signals, strategic politeness, passive aggression, deflection) with three independent annotations per scenario using Plutchik's emotion taxonomy and VAD dimensional ratings.
    \item \textbf{A reusable quality-control pipeline} that combines automated schema validation, statistical outlier detection, agreement analysis, and expert adjudication to ensure annotation quality even when inter-annotator agreement is inherently low.
    \item \textbf{A standardized evaluation protocol} with baselines across 7 LLMs (4 commercial, 3 open-weight) in three prompt modes (zero-shot, chain-of-thought, few-shot), establishing that current models reach only 25\% accuracy on pragmatic emotion inference versus 54\% human majority agreement---and that neither CoT nor few-shot prompting meaningfully improves performance.
\end{enumerate}

\section{Related Work}
\label{sec:related}

\subsection{Pragmatic Reasoning in NLP}

The theoretical apparatus for pragmatic reasoning runs from Gricean maxims \citep{grice1975logic} and speech act theory \citep{searle1975indirect} through relevance theory \citep{sperber1986relevance} to the Rational Speech Act framework \citep{goodman2016pragmatic}. \citet{levinson1983pragmatics} provides a comprehensive treatment of the phenomena we target: indirect speech acts, conversational implicature, and context-dependent interpretation. Each framework makes different commitments about how listeners recover meaning: Grice's cooperative principle assumes shared rationality, relevance theory replaces the maxims with a single cognitive principle of relevance maximization, and the RSA framework \citep{goodman2016pragmatic} formalizes this as recursive Bayesian inference between speaker and listener models. On the computational side, prior work has tackled implicature resolution \citep{jeretic2020natural}, pragmatic inference in dialogue \citep{andreas2016reasoning}, and figurative language understanding \citep{joshi2017automatic}. However, these efforts have largely targeted individual phenomena in isolation: \citet{jeretic2020natural} focus on scalar implicatures and presuppositions, while \citet{andreas2016reasoning} model reference games rather than the socially situated indirect speech that characterizes everyday communication. Recent work has begun evaluating whether LLMs exhibit pragmatic competence: \citet{ruis2023goldilocks} show that fine-tuning strategy significantly affects implicature resolution, and \citet{shapira2024clever} find that apparent theory-of-mind performance in LLMs often relies on shallow heuristics rather than genuine social reasoning. \citet{hu2023fine} provide a fine-grained comparison showing that while LLMs match human performance on conventional implicatures, they fall short on non-conventional and context-dependent inferences---precisely the phenomena CEI targets. A persistent limitation of prior computational approaches is the absence of social context: most benchmarks strip away the speaker-listener relationship, power dynamics, and situational framing that are essential for pragmatic interpretation in practice.

\subsection{Theory of Mind Benchmarks}

Theory of mind (ToM) is closely related to pragmatic reasoning: understanding what a speaker \textit{means} requires modeling what they \textit{believe} and \textit{intend}. Several benchmarks evaluate ToM in LLMs. ToMi \citep{le2019revisiting} tests false-belief understanding in short narratives, while FANToM \citep{kim2023fantom} stress-tests ToM in multi-party conversations. More recently, BigToM \citep{gandhi2023bigtom} evaluates social reasoning through causal templates, Hi-ToM \citep{wu2023hitom} tests higher-order belief reasoning (``Alice thinks Bob thinks...''), and OpenToM \citep{xu2024opentom} provides a comprehensive evaluation covering diverse ToM dimensions. Performance on these benchmarks varies widely: \citet{kosinski2024evaluating} reported that GPT-4 solved 95\% of standard false-belief tasks, but \citet{ullman2023large} showed that trivial alterations to the same tasks---such as making the container transparent---caused performance to collapse, suggesting pattern matching rather than genuine mental-state reasoning. \citet{sap2022neural} further demonstrated that LLMs struggle with social intelligence tasks requiring inference about emotions, social norms, and communicative intent, achieving only 55--65\% accuracy on the SocialIQa benchmark even with chain-of-thought prompting. Event2Mind \citep{rashkin2018event2mind} bridges ToM and emotion by requiring models to infer likely reactions and intents from events, but uses short event descriptions rather than the situated, pragmatically rich utterances we study. These benchmarks focus primarily on belief attribution; CEI complements them by targeting emotional inference from pragmatically complex utterances, where the challenge is not tracking who knows what, but inferring how the speaker \textit{feels} given the social and communicative context.

\subsection{Emotion Recognition Benchmarks}

We situate CEI within the landscape of emotion-annotated datasets. GoEmotions \citep{demszky2020goemotions} provides 58K Reddit comments labeled with 27 emotion categories, achieving a Krippendorff's $\alpha$ of 0.46 across its fine-grained taxonomy. MELD \citep{poria2019meld} and IEMOCAP \citep{busso2008iemocap} offer multimodal emotion recognition in conversations, while CMU-MOSEI \citep{zadeh2018multimodal} extends this to the wild with over 23K sentence-level annotations. EmoBank \citep{buechel2017emobank} annotates 10K sentences for VAD dimensions from both reader and writer perspectives, reporting Pearson correlations of 0.64--0.72 between the two perspectives. SemEval-2018 Task 1 \citep{mohammad2018semeval} established shared-task benchmarks for affect in tweets, using the NRC Emotion Lexicon \citep{mohammad2012semeval} as a foundation. Earlier work on affective text \citep{strapparava2007semeval} used just 1K newspaper headlines with 6 emotion categories, achieving modest agreement ($\kappa \approx 0.28$ for fine-grained labels). \citet{bostan2018analysis} survey these corpora and identify substantial variation in annotation schemes and agreement levels, noting that agreement tends to be lower with finer-grained taxonomies and that $\kappa$ values between 0.20 and 0.40 are common for emotion tasks. CEI differs from these datasets in three respects: we focus specifically on \textit{pragmatically complex} utterances where literal and intended meaning diverge; we annotate the \textit{speaker's} emotion (not the reader's response); and we provide both categorical (Plutchik) and dimensional (VAD) labels per scenario. Unlike GoEmotions and SemEval data, which draw on naturally occurring text where the dominant emotion is typically congruent with surface language, CEI deliberately selects for scenarios where surface cues are misleading---making the annotation task harder but more diagnostic of genuine pragmatic competence.

\subsection{Sarcasm and Indirect Speech}

Sarcasm detection has been studied extensively. SARC \citep{khodak2018large} provides 1M self-labeled Reddit comments and iSarcasm \citep{oprea2020isarcasm} offers 4.5K instances with author-intended labels. Notably, iSarcasm's use of author-intended labels (rather than third-party annotations) revealed that models trained on SARC's self-labeled ``/s'' tags learn a distribution of sarcasm markers that differs substantially from how speakers actually intend sarcasm, highlighting the gap between surface detection and genuine pragmatic understanding. Politeness and indirect speech have received separate attention: the Stanford Politeness Corpus \citep{danescu2013computational} annotates politeness in online requests, building on Brown and Levinson's foundational theory of face-threatening acts \citep{brown1987politeness}. Brown and Levinson's framework posits that speakers use indirect speech to manage face threats, with the degree of indirection scaling with the severity of the face threat and the power distance between interlocutors---a dynamic that our power-relation annotations are designed to capture. \citet{pinker2008logic} formalize this further, arguing that indirect speech serves as a game-theoretic strategy for maintaining plausible deniability in social interactions. Power dynamics in language have been studied by \citet{prabhakaran2012power}, who predict overt displays of power in email threads, and by \citet{bramsen2011extracting}, who extract social power relationships from organizational text. However, these studies treat power as a classification target rather than as a contextual variable that modulates pragmatic interpretation, which is the role it plays in CEI.

\subsection{Annotation Methodology and Agreement}

Our annotation methodology draws on established practice. \citet{artstein2008inter} survey inter-coder agreement measures, including the Fleiss' $\kappa$ \citep{fleiss1971measuring} we use for categorical labels and the ICC \citep{shrout1979intraclass} we use for VAD dimensions. \citet{krippendorff2018content} argues that the threshold for ``acceptable'' agreement depends on the consequences of disagreement and the inherent subjectivity of the coding task. A growing body of work challenges the assumption that low agreement is always problematic: \citet{passonneau2014benefits} argue that annotator disagreement can itself be a valuable signal; \citet{plank2014linguistically} distinguish ``linguistically debatable'' items from those that are ``just plain wrong''; \citet{pavlick2019inherent} demonstrate that human disagreement on NLI is systematic and reproducible; and \citet{uma2021learning} argue that collapsing multi-annotator labels to a single gold standard discards useful information. We adopt this perspective: while we derive majority-vote gold labels for evaluation convenience, we release all individual annotations so that researchers can model the full distribution of human judgments. For dimensional annotation, we follow \citet{warriner2013norms} and \citet{mohammad2018obtaining} in using VAD scales, grounded in the circumplex model of affect \citep{russell1980circumplex,mehrabian1996pleasure} and ANEW norms \citep{bradley1999affective}. Our use of Plutchik's 8 basic emotions \citep{plutchik1980general,plutchik2001nature} alongside VAD ratings provides complementary categorical and dimensional perspectives \citep{buechel2017emobank}.

\subsection{Label Taxonomy Mismatch in LLM Evaluation}

A practical challenge in using fixed emotion taxonomies for LLM evaluation is that models frequently produce out-of-vocabulary (OOV) labels even when explicitly instructed otherwise. \citet{niu2024rethinking} show that GPT-4 systematically diverges from human annotation schemes, and \citet{lian2025ovmer} document 236 distinct labels produced by unconstrained annotators. Several approaches to cross-taxonomy mapping have emerged: GoEmotions \citep{demszky2020goemotions} defines explicit reductions from fine-grained to basic emotions; \citet{buechel2021label} learn label-agnostic emotion embeddings; \citet{lim2024plutchik} show that grounding in Plutchik's wheel structure improves classification; the MER 2024 challenge \citep{lian2024mer} clusters semantically equivalent terms; and the NRC Emotion Lexicon \citep{mohammad2012semeval} provides validated word-to-Plutchik mappings for $\sim$27K terms. \citet{buechel2016corpus} address corpus-level label harmonization across different annotation schemes. These precedents establish that taxonomy harmonization is standard practice in emotion benchmarking; we apply this approach in our evaluation pipeline (Section~\ref{sec:eval}).

\subsection{Positioning of CEI}

CEI differs from all of the above in combining multiple pragmatic subtypes, explicit power relations, and multi-annotator labels with documented agreement, using contextual emotional inference as the evaluation target. Table~\ref{tab:benchmark-comparison} compares CEI against the most closely related pragmatic reasoning benchmarks.

\begin{table}[t]
\centering
\caption{Comparison of CEI with related pragmatic reasoning benchmarks.}
\label{tab:benchmark-comparison}
\small
\begin{tabular}{@{}lcccc@{}}
\toprule
\textbf{Feature} & \textbf{CEI} & \textbf{DiPlomat} & \textbf{PUB} & \textbf{SocialIQa} \\
\midrule
Size & 300 & 4,177 & 28,000 & 38,000 \\
Task & Emotion & PIR + CQA & 14 tasks & MC QA \\
Data source & Expert & AMT & Mixed & Crowdsource \\
Multi-annotator & 3 per item & No & 3 eval. & No \\
Social context & Explicit & Situated & Minimal & Implicit \\
Power relations & 3 types & None & None & None \\
Emotion labels & Plutchik+VAD & None & None & None \\
Pragmatic subtypes & 5 types & 2 types & 14 types & None \\
\bottomrule
\end{tabular}
\end{table}

CEI is smaller than crowd-sourced alternatives but provides features absent from larger benchmarks: explicit power-relation metadata, dual categorical+dimensional emotion annotation, and per-annotator labels with documented agreement across five distinct pragmatic subtypes. DiPlomat \citep{hyun2023diplomat} targets pragmatic reasoning in diplomatic settings but does not annotate emotions; PUB evaluates 14 pragmatic phenomena but provides minimal social context; SocialIQa \citep{sap2019socialiqa} uses a multiple-choice format with single gold answers rather than multi-annotator labels. We follow best practices in dataset documentation \citep{gebru2021datasheets,bender2018data} and reproducibility \citep{pineau2021improving}. To our knowledge, CEI is the first benchmark that jointly evaluates emotion inference across multiple forms of indirect speech, with explicit social context and power relation metadata, while preserving and documenting the full spectrum of human disagreement.

\section{Dataset Design}
\label{sec:design}

Each CEI scenario presents a pragmatically complex communicative exchange designed to test whether a model (or human annotator) can move beyond surface semantics to infer what the speaker actually feels. We describe the structural components below, followed by the annotation task and worked examples from each subtype.

\subsection{Scenario Design}

Every scenario consists of four components: (1) a \textit{context} of 2--4 sentences establishing the social situation; (2) labeled \textit{speaker and listener roles} encoding the interpersonal relationship and power dynamic; (3) an \textit{utterance} produced by the speaker, which serves as the pragmatically ambiguous target; and (4) an \textit{annotation target} requiring the annotator to infer the speaker's primary emotion and provide a Valence-Arousal-Dominance (VAD) profile. This four-part structure ensures that every scenario provides sufficient information for pragmatic disambiguation while keeping the annotation task focused on a single inference: what does the speaker feel?

We organize the 300 scenarios into five pragmatic subtypes (Table~\ref{tab:subtypes}), each corresponding to a well-studied form of indirect speech \citep{levinson1983pragmatics}. Sarcasm and irony involve saying the opposite of what is meant, often with mocking or critical intent \citep{joshi2017automatic}; mixed signals present conflicting verbal and contextual cues that create genuine ambiguity about the speaker's state; strategic politeness masks criticism or negative intent behind surface courtesy, drawing on face-management strategies described by \citet{brown1987politeness}; passive aggression expresses indirect hostility through apparent compliance or withdrawal; and deflection avoids uncomfortable topics through redirection or subject change. We chose these five subtypes because they span a range of pragmatic strategies requiring qualitatively different kinds of contextual reasoning to decode: from recognizing tonal inversion (sarcasm) to detecting emotional avoidance (deflection). Together, they cover the major forms of pragmatic complexity that arise in everyday social interaction.

\begin{table}[t]
\caption{CEI Pragmatic Subtypes with Definitions and Examples}
\label{tab:subtypes}
\centering
\small
\begin{tabular}{@{}p{2.2cm}p{4.5cm}p{5.5cm}@{}}
\toprule
\textbf{Subtype} & \textbf{Definition} & \textbf{Example Utterance} \\
\midrule
Sarcasm/Irony & Saying the opposite of what is meant, often with mocking intent & ``Oh sure, because \textit{that} went well last time.'' \\
\addlinespace
Mixed Signals & Conflicting verbal/contextual cues creating ambiguity & ``I'm fine'' (said with tears) \\
\addlinespace
Strategic Politeness & Surface politeness masking criticism or negative intent & ``That's certainly \textit{one} approach.'' \\
\addlinespace
Passive Aggression & Indirect hostility through apparent compliance & ``No, no, I'll do it myself. Again.'' \\
\addlinespace
Deflection & Avoiding uncomfortable topics through redirection & ``Anyway, how about that weather?'' \\
\bottomrule
\end{tabular}
\end{table}

Scenarios are drawn from four social settings---workplace (meetings, performance discussions), family (parent-child, sibling dynamics), social/friendship (casual gatherings, group outings), and service encounters (customer-staff exchanges)---chosen because indirect speech serves different social functions in each \citep{pinker2008logic}. A polite deflection at work, where career consequences are at stake, operates differently from one at a family dinner, where long-standing relational dynamics shape interpretation. Each scenario is also tagged with one of three power relations between speaker and listener: \textit{peer-to-peer} (N=217, 72\%; colleagues, friends, siblings), \textit{high-to-low} (N=61, 20\%; boss$\rightarrow$employee, parent$\rightarrow$child), and \textit{low-to-high} (N=22, 7\%; employee$\rightarrow$boss, student$\rightarrow$teacher). The peer-heavy distribution mirrors the fact that most everyday communication occurs between equals, but the 83 non-peer scenarios enable preliminary analyses of how power asymmetry modulates pragmatic strategy \citep{prabhakaran2012power}---for instance, whether strategic politeness is more common (or harder to decode) when directed upward in a hierarchy.

\subsection{The Annotation Task}

\begin{figure}[t]
\centering
\begin{tikzpicture}[
    node distance=1.2cm and 2.5cm,
    person/.style={draw, circle, minimum size=1.0cm, font=\small\bfseries, thick},
    context/.style={draw, dashed, rounded corners=8pt, inner sep=12pt, gray!70, thick},
    label/.style={font=\scriptsize, text=black!60},
    arr/.style={-{Stealth[length=3mm]}, thick},
    thought/.style={draw, rounded corners, dashed, font=\scriptsize, text width=2.8cm, align=center, fill=yellow!8},
]
\node[context, minimum width=7.5cm, minimum height=2.8cm] (ctx) {};
\node[label, above] at (ctx.north) {\textit{Social Context}};

\node[person, fill=blue!15] (speaker) at (-1.8, 0) {S};
\node[label, below=2pt] at (speaker.south) {Speaker};
\node[person, fill=green!15] (listener) at (1.8, 0) {L};
\node[label, below=2pt] at (listener.south) {Listener};

\draw[arr] (speaker) -- (listener) node[midway, above, font=\small\itshape] {``utterance''};

\node[label] at (0, -0.45) {power relation};

\node[person, fill=orange!20] (annotator) at (0, -3.2) {A};
\node[label, below=2pt] at (annotator.south) {Annotator};

\draw[arr, dashed, gray] (annotator) -- (-1.2, -1.4);
\draw[arr, dashed, gray] (annotator) -- (1.2, -1.4);

\node[thought, right=0.8cm] at (annotator.east) {Infer: Plutchik emotion\\+ VAD ratings\\+ confidence};
\end{tikzpicture}
\caption{The CEI annotation task. A Speaker (S) directs an utterance to a Listener (L) within a shared social context with an explicit power relation. An external Annotator (A) observes the exchange and infers the speaker's primary emotion (from Plutchik's 8 categories), VAD ratings, and their own confidence.}
\label{fig:annotation-task}
\end{figure}
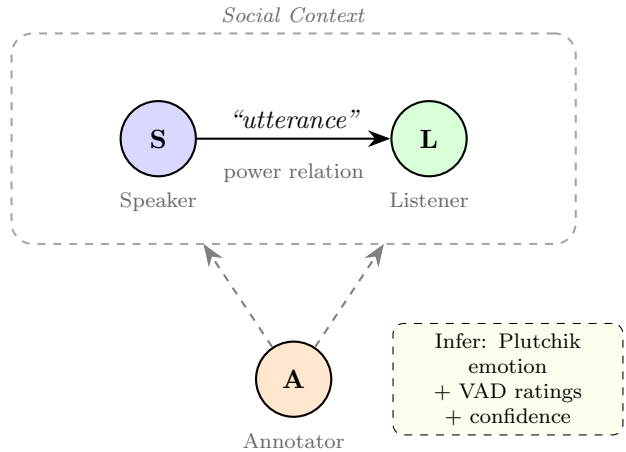

Figure~\ref{fig:annotation-task} illustrates the inference problem that each scenario poses. A speaker produces an utterance directed at a listener within a shared social context; an external annotator observes this exchange and must infer the speaker's primary emotion from Plutchik's 8 basic emotions \citep{plutchik1980general} and rate the speaker's affective state along three VAD dimensions \citep{mehrabian1996pleasure,russell1980circumplex}. The key challenge is that the utterance's literal content may be misleading: a speaker saying ``That's certainly \textit{one} approach'' may feel anger, disgust, or even amusement, and the annotator must use the context, roles, and power dynamic to disambiguate. This setup mirrors the real-world challenge of pragmatic understanding: an observer must integrate multiple information sources---situational context, social roles, relational history, and communicative norms---to recover the intended emotional meaning from what is literally said.

\subsection{Worked Examples}

We present five scenarios---one per pragmatic subtype---drawn directly from the released data. These illustrate the dataset's structure, the range of annotator responses, and the interpretive challenges each subtype poses.

\paragraph{Sarcasm/irony (unanimous agreement).} \textit{Context:} After a programmer accidentally pushed a code update that reverted most of the day's progress, a teammate sighs and addresses the small group of peers who worked on it together. \textit{Speaker:} teammate (peer). \textit{Utterance:} ``Well, that went\ldots\ great.'' \textit{Annotations:} sadness, sadness, sadness $\rightarrow$ \textit{Gold:} sadness. All three annotators converged on sadness rather than anger, reading the sarcastic understatement as an expression of deflated disappointment rather than hostility. VAD ratings were consistent: mildly unpleasant valence, low arousal, moderate control. This is one of the 14\% of scenarios where all annotators agree, likely because the sarcasm is transparent and the emotional register (resigned disappointment) is unambiguous.

\paragraph{Passive aggression (three-way split).} \textit{Context:} You and your sibling usually trade cooking and chores. Lately they cancel plans and expect you to cover evenings. Tonight they say they need to head out with friends right before dinner. \textit{Speaker:} you (peer). \textit{Utterance:} ``Oh, it's fine, go ahead and go; I can make dinner again, don't worry about me.'' \textit{Annotations:} sadness, surprise, joy $\rightarrow$ \textit{Gold:} sadness. This three-way split exemplifies the difficulty of passive-aggressive utterances: the surface compliance (``it's fine'') can be read as resigned sadness, incredulous surprise at the sibling's behavior, or even sarcastic joy. The meta-annotator selected sadness based on the accumulating resentment described in the context. VAD ratings diverged across all three dimensions, underscoring the genuine ambiguity.

\paragraph{Deflection (majority agreement).} \textit{Context:} In the office kitchen, one colleague brings up that Jamie's slides skipped several key data points and asks if Jamie can explain. \textit{Speaker:} Jamie (peer). \textit{Utterance:} ``No worries, I've brought donuts for everyone, let's all grab one! Also, who's up for moving the Friday coffee run to earlier?'' \textit{Annotations:} surprise, surprise, anger $\rightarrow$ \textit{Gold:} surprise. Two annotators read the deflection as revealing discomfort (surprise at being called out); one interpreted underlying frustration (anger). VAD ratings converged on mildly unpleasant valence with moderate control. The rapid subject change is a classic deflection marker, but annotators disagreed on whether the speaker's underlying state is anxious avoidance or irritated dismissal.

\section{Annotation Process}
\label{sec:annotation}

\subsection{Annotators and Protocol}

We recruited 15 undergraduate students from an interdisciplinary course (IPHS 391) at Kenyon College, all fluent English speakers. Each annotator received two 20-minute in-class sessions: the first covered the theoretical framework (pragmatic subtypes, Plutchik's emotion model, VAD dimensions) with worked examples, and the second was a live labeling demonstration using the Label Studio interface. In addition, annotators were given both a concise 1-page annotation cheatsheet and a 3-page written guideline document (Appendix~\ref{app:guidelines}) covering definitions and examples for all five pragmatic subtypes, Plutchik's 8 basic emotions with prototypical utterances, VAD scales with anchored examples at each rating point, and the distinction between difficulty (how hard the item is to interpret) and confidence (how sure the annotator feels). Annotators reviewed the guidelines independently before starting. No formal practice round with calibration feedback was conducted, a limitation we acknowledge in \S\ref{sec:limitations}. However, the free-text explanation requirement (2--4 sentences per scenario describing reasoning and alternative interpretations considered) served as a de facto engagement check: annotators who wrote thoughtful, specific explanations demonstrated task understanding even without a formal calibration step. Our quality control pipeline (\S\ref{sec:annotation}) verified that only 2.6\% of annotations were flagged for potential quality issues (timing anomalies or self-contradictions), and no annotator exceeded a 5\% flag rate.

Each annotator was assigned all 60 scenarios of a single subtype (3 annotators per subtype, 15 total across 5 subtypes). This within-subtype design ensures that per-subtype $\kappa$ values reflect a consistent rater set across all 60 items, enabling valid comparisons of agreement difficulty across subtypes, though it precludes assessment of cross-subtype rater consistency. The annotation was completed as a graded course assignment over 12 days in November--December 2025, with each annotator completing their 60 scenarios in a single session of roughly one hour. Beyond the categorical and dimensional labels, annotators wrote a 2--4 sentence free-text explanation for each scenario describing their reasoning and any alternative interpretations they considered. These explanations are not released (to protect annotator privacy) but informed the meta-annotator's adjudication decisions. Annotators whose work met internal consistency thresholds were offered co-authorship.

\begin{figure}[t]
    \centering
    \includegraphics[width=\columnwidth]{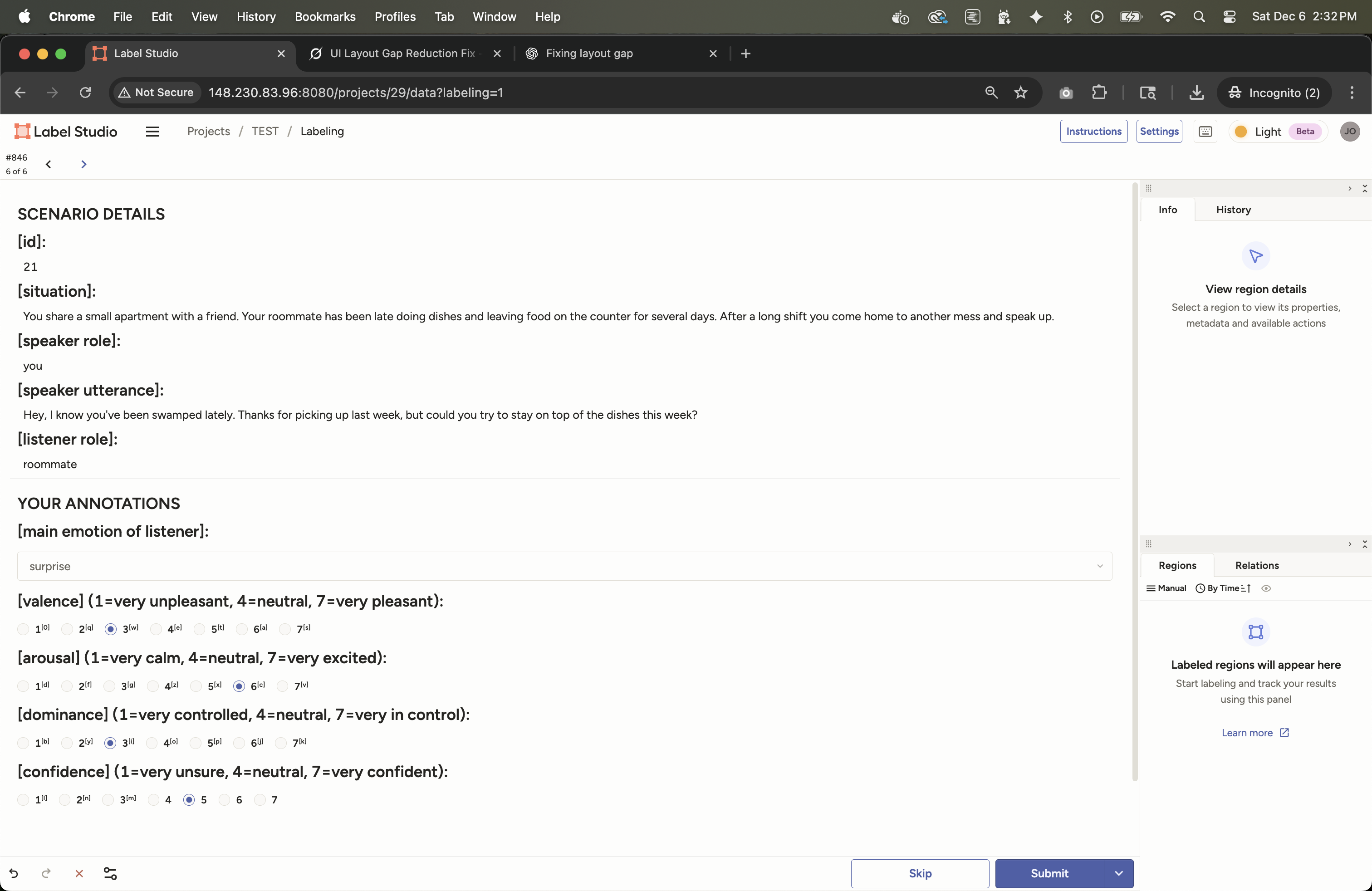}
    \caption{Label Studio annotation interface. Annotators viewed the situational context, speaker/listener roles, and utterance, then selected a primary emotion from Plutchik's 8 categories and provided Valence-Arousal-Dominance ratings on 7-point scales.}
    \label{fig:labelstudio}
\end{figure}

For each scenario, annotators provided three types of judgments: (1) a \textit{primary emotion} selected from Plutchik's 8 basic emotions (joy, trust, fear, surprise, sadness, disgust, anger, anticipation); (2) \textit{VAD ratings} for valence, arousal, and dominance on 7-point scales mapped to $[-1.0, +1.0]$; and (3) a \textit{confidence} rating on the same scale. Annotations were collected via Label Studio \citep{labelstudio2020} (Figure~\ref{fig:labelstudio}), which presented the situational context, speaker/listener roles, and utterance in a structured interface. Annotators spent roughly 1 minute per scenario (median), consistent with the focused nature of the task: reading a short context, considering the pragmatic dynamics, and selecting an emotion category is cognitively demanding but not time-intensive.

\subsection{Quality Control Pipeline}

\begin{figure}[t]
\centering
\begin{tikzpicture}[
    node distance=0.55cm,
    stage/.style={draw, rounded corners, minimum width=5.8cm, minimum height=0.7cm,
                  font=\small, align=center, fill=gray!8},
    data/.style={draw, rounded corners, minimum width=3.2cm, minimum height=0.6cm,
                 font=\small\bfseries, fill=white},
    arr/.style={-{Stealth[length=2.5mm]}, thick},
    lbl/.style={font=\scriptsize, text=black!70},
]
\node[data] (raw) {900 annotations};
\node[stage, below=of raw] (l1) {Level~1: Schema validation};
\node[stage, below=of l1] (l2) {Level~2: Statistical consistency};
\node[stage, below=of l2] (l3) {Level~3: Agreement analysis};
\node[stage, below=of l3] (l4) {Level~4: Expert adjudication};
\node[data, below=of l4] (gold) {300 gold labels};

\draw[arr] (raw) -- (l1);
\draw[arr] (l1) -- (l2);
\draw[arr] (l2) -- (l3) node[lbl, midway, right=2pt] {2.6\% flagged};
\draw[arr] (l3) -- (l4) node[lbl, midway, right=2pt] {$\kappa$ + VAD divergence};
\draw[arr] (l4) -- (gold) node[lbl, midway, right=2pt] {15.7\% adjudicated};
\end{tikzpicture}
\caption{Quality control pipeline. Annotations pass through four stages: schema checks, statistical outlier detection, inter-annotator agreement analysis, and expert adjudication of flagged items.}
\label{fig:qa_pipeline}
\end{figure}
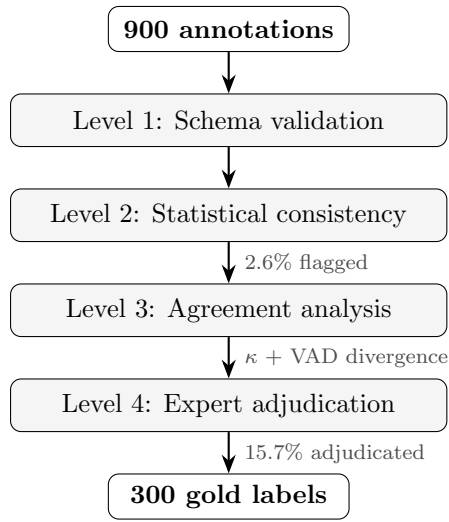

We implement quality control in four stages (Figure~\ref{fig:qa_pipeline}), designed to distinguish genuine annotation errors from the legitimate disagreement that is expected on pragmatically complex items.

\paragraph{Level~1: Schema validation.} We check each annotation record for JSON structure, required fields (emotion, VAD ratings, confidence), and valid enum values. Records with missing or malformed entries are rejected before any analysis. This stage catches data-entry errors and export artifacts rather than substantive annotation problems.

\paragraph{Level~2: Statistical consistency.} We apply three automated checks to flag suspect annotations. \textit{Straight-lining} detects annotators who assign the same emotion to more than 80\% of their scenarios, which would suggest disengagement rather than genuine judgment. \textit{Timing outliers} use Median Absolute Deviation (MAD) with a modified Z-score threshold of 2.5; responses under 3 seconds are flagged as implausibly fast (insufficient time to read the context), and those over 600 seconds as disengaged. \textit{Self-contradiction} identifies cases where an annotator pairs a positive emotion (e.g., joy) with a strongly negative valence rating, or vice versa, since such pairings suggest careless responding rather than a defensible interpretation. Across all 900 annotations, 2.6\% were flagged at this stage; none met the straight-lining threshold, and no annotator was excluded.

\paragraph{Level~3: Agreement analysis.} We compute Fleiss' $\kappa$ per subtype with 2,000 bootstrap resamples for 95\% confidence intervals. Scenarios with complete three-way disagreement (all three annotators choose different emotions) and those with high VAD divergence (pairwise range $>1.5$ on the 7-point scale) are flagged for Level~4 review. We emphasize that low $\kappa$ at this stage is \textit{not} treated as evidence of poor annotation quality; rather, it identifies scenarios that warrant expert review to determine whether the disagreement reflects genuine ambiguity or an annotation error.

\paragraph{Level~4: Expert adjudication.} A trained meta-annotator reviewed all flagged scenarios (15.7\% of the dataset, 47 of 300). For scenarios where two of three annotators agreed, the majority label became the gold standard. For three-way splits (31.3\% of scenarios), the meta-annotator applied two heuristics: first, whether annotators agreed on valence polarity despite disagreeing on the emotion category (e.g., all chose negative emotions); second, whether one annotator's response appeared to reflect a misreading of the scenario context. The meta-annotator documented reasoning for each adjudicated label in the ``Notes'' column of the released CSVs; 72 scenarios carry such notes across three subtypes (mixed signals, deflection, strategic politeness). In cases where no principled resolution was possible, the meta-annotator chose the label most consistent with the VAD ratings and flagged the scenario as genuinely ambiguous. This approach follows the recommendation of \citet{passonneau2014benefits} that adjudication should preserve rather than suppress legitimate disagreement.

\section{Dataset Statistics}
\label{sec:stats}

\subsection{Overall Distribution}

We summarize the dataset composition in Table~\ref{tab:distribution}.

\begin{table}[t]
\caption{CEI Dataset Distribution}
\label{tab:distribution}
\centering
\small
\begin{tabular}{@{}lr@{}}
\toprule
\textbf{Attribute} & \textbf{Value} \\
\midrule
Total scenarios & 300 \\
Annotations per scenario & 3 \\
Total annotations & 900 \\
Pragmatic subtypes & 5 \\
Scenarios per subtype & 60 \\
Power relations & 3 \\
Social contexts & Workplace, family, social, service \\
\bottomrule
\end{tabular}
\end{table}

Label distribution is imbalanced. Surprise (23\%), sadness (21\%), and anger (17\%) account for over 60\% of gold-standard labels, while disgust (4\%) and joy (7\%) are rare. The remaining emotions (fear, anticipation, and trust) each appear in roughly 9\% of scenarios. This skew partly reflects the dataset's focus on pragmatically complex utterances, which tend to involve negative or ambiguous affect. The imbalance should be kept in mind when interpreting macro-F1 scores (\S\ref{sec:eval}), as rare classes pull the average down.

\subsection{Inter-Annotator Agreement}

We report Fleiss' $\kappa$ by subtype in Table~\ref{tab:agreement}. Because we assigned each annotator to one subtype (\S\ref{sec:annotation}), per-subtype values reflect the same 3 raters across all 60 scenarios. The overall $\kappa$ pools all 300 scenarios. Fair agreement ($\kappa = 0.21$) is expected: pragmatic utterances support multiple valid readings, and even attentive annotators frequently disagree on the emotional valence of indirect speech.

\begin{table}[t]
\caption{Inter-Annotator Agreement by Subtype (Fleiss' $\kappa$)}
\label{tab:agreement}
\centering
\small
\begin{tabular}{@{}lccc@{}}
\toprule
\textbf{Subtype} & \textbf{$\kappa$} & \textbf{95\% CI} & \textbf{Interpretation} \\
\midrule
Sarcasm/Irony & 0.25 & [0.15, 0.34] & Fair \\
Passive Aggression & 0.22 & [0.13, 0.30] & Fair \\
Strategic Politeness & 0.20 & [0.10, 0.29] & Slight \\
Mixed Signals & 0.16 & [0.07, 0.25] & Slight \\
Deflection & 0.06 & [$-$0.02, 0.14] & Slight \\
\midrule
\textbf{Overall} & \textbf{0.21} & [0.16, 0.25] & Fair \\
\bottomrule
\end{tabular}
\end{table}

\subsection{Human Performance}

Annotators agree less often than not. We measure human ``accuracy'' as agreement with ground truth (majority vote):

\begin{itemize}[leftmargin=*,noitemsep]
    \item \textbf{Mean annotator agreement:} 61\%
    \item \textbf{Unanimous agreement:} 14.3\% of scenarios (43/300)
    \item \textbf{Majority agreement:} 54.3\% of scenarios (163/300)
    \item \textbf{Three-way split:} 31.3\% of scenarios (94/300)
\end{itemize}

\noindent Individual annotator agreement with gold ranges from 45.0\% to 81.7\% (mean 60.9\%, $N$=15). Critically, no single annotator drives the low overall $\kappa$: per-subtype agreement rates are sarcasm/irony 64.4\%, passive aggression 64.4\%, strategic politeness 62.2\%, mixed signals 60.9\%, and deflection 52.2\%. Even the highest-performing annotator (81.7\% on sarcasm/irony) is well below ceiling, confirming that the task is genuinely difficult for humans. Per-annotator quality metrics (flag rates, timing, gold agreement) are reported in Appendix~\ref{app:logistics}.

Figure~\ref{fig:difficulty} shows the difficulty distribution overall and by subtype.

\begin{figure}[t]
    \centering
    \includegraphics[width=\columnwidth]{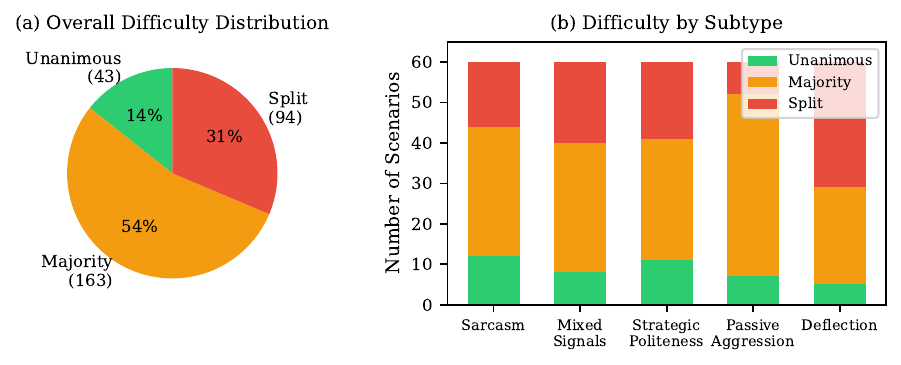}
    \caption{Scenario difficulty distribution based on annotator agreement. (a) Overall: 14\% unanimous, 54\% majority, 31\% split. (b) By subtype: sarcasm/irony shows highest agreement; deflection shows lowest.}
    \label{fig:difficulty}
\end{figure}

\subsection{Emotion Confusion Patterns}

\begin{figure}[t]
    \centering
    \includegraphics[width=0.85\columnwidth]{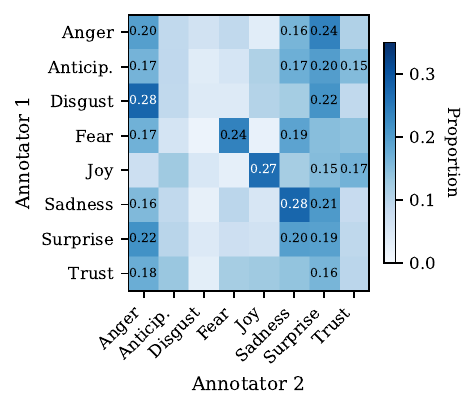}
    \caption{Human annotator confusion matrix for Plutchik's 8 emotions. Values show row-normalized proportions (how often Annotator 1's choice co-occurs with Annotator 2's). Diagonal represents agreement. Anger-surprise and sadness-surprise show highest confusion.}
    \label{fig:confusion}
\end{figure}

Certain emotion pairs are systematically confused. Figure~\ref{fig:confusion} shows the 8$\times$8 confusion matrix aggregated across all annotations. We find three dominant confusion pairs:
\begin{itemize}[leftmargin=*,noitemsep]
    \item Anger $\leftrightarrow$ Surprise (64 pairs)
    \item Sadness $\leftrightarrow$ Surprise (58 pairs)
    \item Anger $\leftrightarrow$ Sadness (44 pairs)
\end{itemize}

All three pairs involve negative emotions with overlapping valence profiles, consistent with known difficulty in distinguishing them without richer context \citep{russell1980circumplex}.

\subsection{Disagreement Analysis}
\label{sec:disagreement}

We present quantitative evidence that annotator disagreement on CEI reflects systematic ambiguity rather than random noise. Three analyses support this claim.

\paragraph{Valence coherence in split scenarios.} Among the 94 three-way splits (scenarios where all 3 annotators chose different emotions), 40.4\% (38/94) nonetheless showed complete agreement on valence polarity: 34 scenarios had all-negative valence, 3 all-positive, and 1 all-neutral. This means that even when annotators completely disagree on the emotion category, they often converge on the overall affective direction---overwhelmingly negative, consistent with the pragmatically complex content of CEI scenarios. Random disagreement would not produce this pattern.

\paragraph{Wheel-adjacency structure.} Across all 608 pairwise disagreements (pairs of annotators assigning different emotions on the same scenario), 31.4\% (191/608) involved Plutchik-adjacent emotions, compared to 28.6\% expected by chance (each emotion has 2 of 7 possible neighbors). The adjacency rate varies meaningfully by subtype: deflection-misdirection shows the highest rate (39.7\%) and passive aggression the lowest (17.5\%). The dominant adjacent confusion pairs---sadness$\leftrightarrow$surprise (58), anger$\leftrightarrow$anticipation (25), fear$\leftrightarrow$surprise (22), joy$\leftrightarrow$trust (22)---reflect psychologically plausible co-activations rather than random labeling.

\paragraph{Comparison with related benchmarks.} CEI's overall $\kappa = 0.21$ is comparable to agreement rates observed in other complex emotion and pragmatic inference tasks. SemEval-2007 Affective Text reports $\kappa \approx 0.28$ for fine-grained emotion classification; GoEmotions \citep{demszky2020goemotions} reports Krippendorff's $\alpha = 0.46$ for 27-category annotation; and sentiment analysis tasks typically achieve $\kappa > 0.70$. EmoBench \citep{liu2024emobench} reports $\kappa = 0.852$ but uses a binary emotion recognition format (yes/no) on decontextualized descriptions, a substantially simpler task than inferring the speaker's emotional state from pragmatically complex indirect speech. Agreement scales inversely with task complexity and label-set granularity, and CEI targets the hardest segment of this spectrum.

\subsection{Linguistic Feature Analysis}

\begin{figure}[t]
    \centering
    \includegraphics[width=\columnwidth]{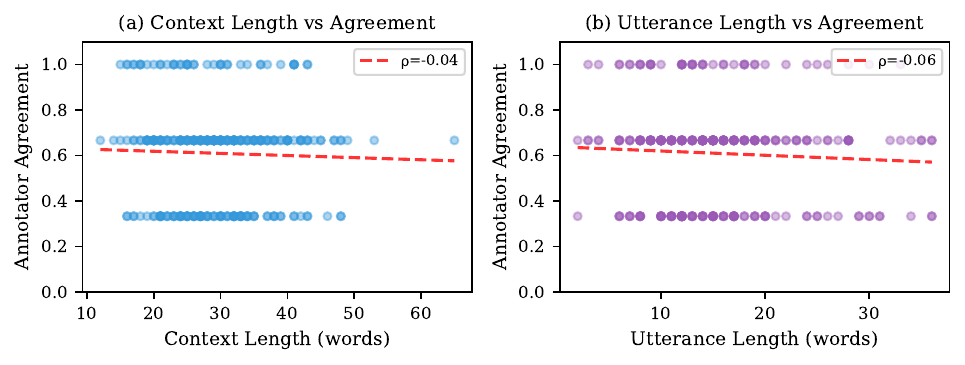}
    \caption{Linguistic features vs. annotator agreement (Spearman $\rho$). (a) Context length shows no correlation with agreement ($\rho=-0.04$). (b) Utterance length similarly uncorrelated ($\rho=-0.06$). Difficulty stems from pragmatic ambiguity, not text complexity.}
    \label{fig:linguistic}
\end{figure}

Length does not predict difficulty. We examine whether scenario length predicts annotation difficulty using Spearman rank correlations (Figure~\ref{fig:linguistic}). Neither context length ($\rho=-0.04$) nor utterance length ($\rho=-0.06$) predicts agreement, suggesting that difficulty comes from pragmatic complexity rather than surface features like verbosity.

\subsection{VAD Dimensional Analysis}

We also collect dimensional affect ratings. Each annotation includes Valence-Arousal-Dominance (VAD) ratings; we report ICC(2,1) values for inter-annotator consistency on each dimension in Table~\ref{tab:vad-icc}.

\begin{table}[t]
\centering
\caption{Inter-annotator consistency on VAD dimensions (ICC(2,1)).}
\label{tab:vad-icc}
\small
\begin{tabular}{@{}lcccccc@{}}
\toprule
\textbf{Dimension} & \textbf{Overall} & \textbf{Sarc.} & \textbf{Mixed} & \textbf{PA} & \textbf{Strat.} & \textbf{Defl.} \\
\midrule
Valence & 0.40 & 0.47 & 0.40 & 0.51 & 0.39 & 0.25 \\
Arousal & 0.20 & 0.26 & 0.17 & 0.22 & $-$0.03 & 0.14 \\
Dominance & 0.21 & 0.38 & 0.26 & 0.28 & $-$0.02 & 0.02 \\
\bottomrule
\end{tabular}
\end{table}

Valence is most reliable. We find the strongest agreement on valence (ICC=0.40), consistent with evidence that affective valence is the most reliably perceived dimension \citep{russell1980circumplex,warriner2013norms}. Arousal and dominance are lower (ICC$\approx$0.20) and vary considerably across subtypes. Strategic politeness and deflection show near-zero ICC on arousal and dominance; these subtypes seem to create particular ambiguity about how intense or in-control the speaker feels.

We use the VAD ratings as a consistency check on the categorical labels: positive emotions (joy: $\bar{V}=+0.48$; trust: $\bar{V}=+0.17$) get positive valence, negative emotions (fear: $\bar{V}=-0.56$; sadness: $\bar{V}=-0.54$; anger: $\bar{V}=-0.51$) get negative valence, and this holds across all subtypes. This consistency suggests that even when annotators disagree on the emotion category, their underlying affective judgments are coherent. We note that VAD prediction could serve as a complementary continuous evaluation target in future work.

\section{Evaluation Framework}
\label{sec:eval}

The CEI evaluation task is straightforward: models receive the context, speaker/listener roles, and utterance, and must predict the speaker's primary emotional state from Plutchik's 8 basic emotions. We recommend reporting accuracy (exact match) and macro-F1 (averaged across the 8 classes), with breakdowns by subtype and power relation. Subtype-level results are particularly informative because they reveal whether a model fails uniformly or has systematic blind spots (e.g., handling sarcasm but not deflection). We provide zero-shot and chain-of-thought prompt templates in Appendix~\ref{app:prompts}.

Despite explicit instructions to choose from Plutchik's 8 emotions, models frequently produce out-of-vocabulary (OOV) labels. In the zero-shot condition, across 2,100 responses (7 models $\times$ 300 scenarios), 46 (2.2\%) contained emotions outside the prescribed set---consistent with recent findings that LLMs systematically diverge from constrained label sets \citep{niu2024rethinking}. The rate varied by model: GPT-5-mini produced the most OOV responses (23/300, 7.7\%), while Llama-3.1-70B and Grok-4.1-Fast produced the fewest (2/300 each). Rather than discard these responses, we apply taxonomy harmonization following established practice in emotion benchmarking \citep{lian2024mer,demszky2020goemotions}. Each OOV label is mapped to its nearest Plutchik emotion based on wheel adjacency and the NRC Emotion Lexicon \citep{mohammad2012semeval}: for instance, ``frustration'' maps to anger (lower-intensity variant on the same petal), ``gratitude'' maps to joy (positive valence cluster), and ``avoidance'' maps to fear (threat-response cluster). Table~\ref{tab:oov-mapping} lists all 19 distinct OOV labels observed in practice and their mappings. Of the 46 OOV responses, 44 were cleanly mappable to Plutchik emotions; the remaining 2 produced garbled text and are scored as incorrect predictions. The full mapping table and salvage code are included in the released repository.

\begin{table}[t]
\centering
\caption{Out-of-vocabulary emotion labels produced by baseline models, mapped to nearest Plutchik emotion via wheel adjacency and the NRC Emotion Lexicon. $n$ = number of occurrences across all 2,100 model responses.}
\label{tab:oov-mapping}
\small
\begin{tabular}{@{}llr@{}}
\toprule
\textbf{OOV Label} & \textbf{Plutchik Mapping} & \textbf{$n$} \\
\midrule
sarcasm        & disgust       & 10 \\
pride          & joy           & 4 \\
gratitude      & joy           & 4 \\
disappointment & sadness       & 3 \\
relief         & joy           & 3 \\
guilt          & sadness       & 2 \\
amusement      & joy           & 2 \\
concern        & trust         & 2 \\
reassurance    & trust         & 2 \\
frustration    & anger         & 2 \\
defiance       & anger         & 2 \\
embarrassment  & fear          & 1 \\
evasion        & fear          & 1 \\
curiosity      & anticipation  & 1 \\
playful        & joy           & 1 \\
avoidance      & fear          & 1 \\
defense        & fear          & 1 \\
resignation    & sadness       & 1 \\
satisfaction   & joy           & 1 \\
\midrule
\textbf{Total} &               & \textbf{44} \\
\bottomrule
\end{tabular}
\end{table}

We primarily evaluate in the zero-shot setting because it tests a model's intrinsic pragmatic competence without task-specific calibration. To verify that the low performance is not an artifact of the zero-shot format, we additionally evaluate all models with chain-of-thought (CoT) and 3-shot prompting (Table~\ref{tab:prompt-modes}). CoT prompting asks models to reason through five steps (literal meaning, contextual cues, pragmatic interpretation, internal state, emotion label) before answering; 3-shot prompting provides three worked examples spanning different subtypes. Neither strategy meaningfully improves performance: mean accuracy is 20.0\% (zero-shot), 20.2\% (CoT), and 18.9\% (few-shot). The best single result across all modes is 25.0\% (Llama-3.1-70B zero-shot and Qwen2.5-7B CoT). These results confirm that the difficulty of CEI reflects a genuine gap in pragmatic competence rather than a prompting artifact.

Prompt mode also affects \textit{response format compliance}, even when it does not affect accuracy. Parse failure rates (responses that could not be extracted as valid JSON with an emotion label) averaged 11.6\% in zero-shot, 8.1\% in CoT, and 9.7\% in few-shot mode. The effect varies sharply by model: GPT-5-mini dropped from 27.7\% parse failures (zero-shot) to 0\% (CoT), and Gemini~2.5~Flash from 37.3\% to 14.0\%. Llama-3.1-70B produced zero parse failures in all three modes. CoT's structured format appears to regularize output even without improving classification accuracy. All parse failures are scored as incorrect predictions; the accuracy values reported throughout this paper reflect this penalty.

\begin{table}[t]
\centering
\caption{Accuracy by prompt mode across all 7 baseline models. Neither chain-of-thought nor few-shot prompting meaningfully improves performance over zero-shot, confirming that CEI difficulty reflects genuine pragmatic reasoning gaps. Best per model is \textbf{bolded}.}
\label{tab:prompt-modes}
\small
\begin{tabular}{@{}lccc@{}}
\toprule
\textbf{Model} & \textbf{Zero-shot} & \textbf{CoT} & \textbf{Few-shot} \\
\midrule
Llama-3.1-70B & \textbf{0.250} & 0.220 & 0.213 \\
DeepSeek-V3 & \textbf{0.220} & 0.193 & 0.197 \\
Grok-4.1-Fast & \textbf{0.213} & 0.210 & 0.203 \\
Qwen2.5-7B & 0.207 & \textbf{0.250} & 0.203 \\
Claude Sonnet 4.5 & \textbf{0.193} & 0.150 & 0.180 \\
GPT-5-mini & 0.187 & \textbf{0.243} & 0.140 \\
Gemini 2.5 Flash & 0.130 & 0.147 & \textbf{0.183} \\
\midrule
\textbf{Mean} & \textbf{0.200} & \textbf{0.202} & 0.189 \\
\bottomrule
\end{tabular}
\end{table}

We evaluate 7 models (4 commercial APIs, 3 open-weight) with temperature 0.\footnote{All inference uses temperature 0 (greedy decoding), making predictions deterministic across runs. The 95\% CIs in Table~\ref{tab:baselines} are bootstrap percentile intervals (10{,}000 resamples over the 300 test scenarios, seed 42) and capture sampling uncertainty---how much accuracy might shift given a different draw of scenarios from the same distribution.} Table~\ref{tab:baselines} reports overall results on the full 300 scenarios after OOV harmonization. The best model reaches 25.0\% accuracy (95\% CI [0.200, 0.300]), roughly double the 12.5\% random baseline, but well below human majority agreement (54\%). All seven model CIs fall entirely below 0.300, confirming that the human--model gap is not an artifact of sampling noise. Macro-F1 scores range from 0.12 to 0.23, indicating that even the best models perform unevenly across emotion classes. This gap between random performance and the model ceiling (about 12.5 percentage points) is small relative to the gap between models and humans (about 29 percentage points), suggesting that current LLMs capture some pragmatic signal but miss most of it. The wide overlap among model CIs also means we cannot confidently rank individual models; the performance differences are within sampling uncertainty for $n=300$. The low ceiling is consistent with findings from related benchmarks: \citet{sap2022neural} report 55--65\% accuracy on SocialIQa with chain-of-thought prompting, and \citet{hu2023fine} show that LLMs fall short on non-conventional inferences.

\begin{table}[t]
\centering
\caption{Zero-shot baseline performance on CEI speaker emotion prediction (300 scenarios, 8-class Plutchik classification, after OOV harmonization). Random baseline accuracy is 12.5\%. 95\% CIs are bootstrap percentile intervals (10{,}000 resamples over scenarios). Best result per column is \textbf{bolded}.}
\label{tab:baselines}
\small
\begin{tabular}{@{}lccc@{}}
\toprule
\textbf{Model} & \textbf{Accuracy} & \textbf{95\% CI} & \textbf{Macro-F1} \\
\midrule
Llama-3.1-70B & \textbf{0.250} & [0.200, 0.300] & \textbf{0.225} \\
DeepSeek-V3 & 0.220 & [0.173, 0.267] & 0.179 \\
Grok-4.1-Fast & 0.213 & [0.170, 0.260] & 0.194 \\
Qwen2.5-7B & 0.207 & [0.163, 0.253] & 0.152 \\
Claude Sonnet 4.5 & 0.193 & [0.150, 0.237] & 0.177 \\
GPT-5-mini & 0.187 & [0.143, 0.230] & 0.192 \\
Gemini 2.5 Flash & 0.130 & [0.093, 0.170] & 0.121 \\
\bottomrule
\end{tabular}
\end{table}

Performance varies substantially by subtype (Table~\ref{tab:baselines-subtype}). Averaging across all 7 models, accuracy is highest on strategic politeness (25.5\%) and mixed signals (22.1\%), and lowest on sarcasm/irony (15.2\%) and passive aggression (16.0\%). The sarcasm result is particularly striking: sarcasm is the subtype where humans agree most ($\kappa=0.25$) yet models perform worst, suggesting that the contextual cues humans rely on for sarcasm detection---tonal incongruity, shared expectations about what ``should'' be said---are not yet accessible to current LLMs through text alone. By contrast, deflection shows the lowest human agreement ($\kappa=0.06$) but middling model performance (21.2\%), suggesting that models and humans find different subtypes difficult. This dissociation between human and model difficulty is itself a finding: it implies that model errors do not simply mirror human uncertainty, but reflect qualitatively different processing failures.

\begin{table}[t]
\centering
\caption{Per-subtype zero-shot accuracy for all baseline models. Best result per column is \textbf{bolded}. Sarc.~= Sarcasm/Irony, PA~= Passive Aggression, Strat.~= Strategic Politeness, Mixed~= Mixed Signals, Defl.~= Deflection.}
\label{tab:baselines-subtype}
\small
\begin{tabular}{@{}lccccc@{}}
\toprule
\textbf{Model} & \textbf{Sarc.} & \textbf{PA} & \textbf{Strat.} & \textbf{Mixed} & \textbf{Defl.} \\
\midrule
Llama-3.1-70B & 0.183 & 0.183 & 0.283 & \textbf{0.333} & 0.267 \\
DeepSeek-V3 & 0.150 & 0.183 & 0.283 & 0.233 & 0.250 \\
Grok-4.1-Fast & 0.183 & 0.183 & 0.250 & 0.217 & 0.233 \\
Qwen2.5-7B & \textbf{0.183} & 0.133 & \textbf{0.333} & 0.150 & 0.233 \\
Claude Sonnet 4.5 & 0.117 & \textbf{0.200} & 0.217 & 0.283 & 0.150 \\
GPT-5-mini & 0.117 & 0.150 & 0.267 & 0.200 & 0.200 \\
Gemini 2.5 Flash & 0.133 & 0.083 & 0.150 & 0.133 & 0.150 \\
\midrule
\textbf{Model avg.} & 0.152 & 0.160 & 0.255 & 0.221 & 0.212 \\
\textbf{Human $\kappa$} & 0.25 & 0.22 & 0.20 & 0.16 & 0.06 \\
\bottomrule
\end{tabular}
\end{table}

\begin{figure}[t]
    \centering
    \includegraphics[width=\columnwidth]{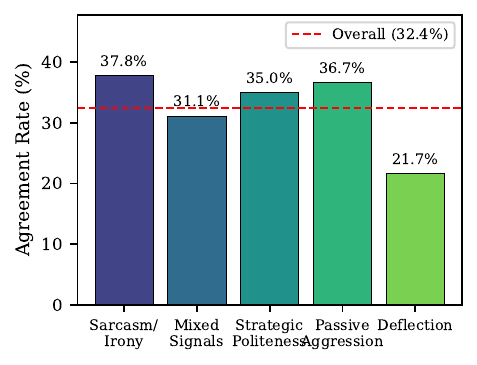}
    \caption{Agreement patterns by pragmatic subtype. Each bar shows the proportion of scenarios with unanimous (all 3 agree), majority (2 of 3), or split (all disagree) annotations. Sarcasm/irony shows the highest unanimity; deflection shows the most three-way splits, consistent with its near-chance $\kappa$ (0.06).}
    \label{fig:agreement-subtype}
\end{figure}

\begin{figure}[t]
    \centering
    \includegraphics[width=\columnwidth]{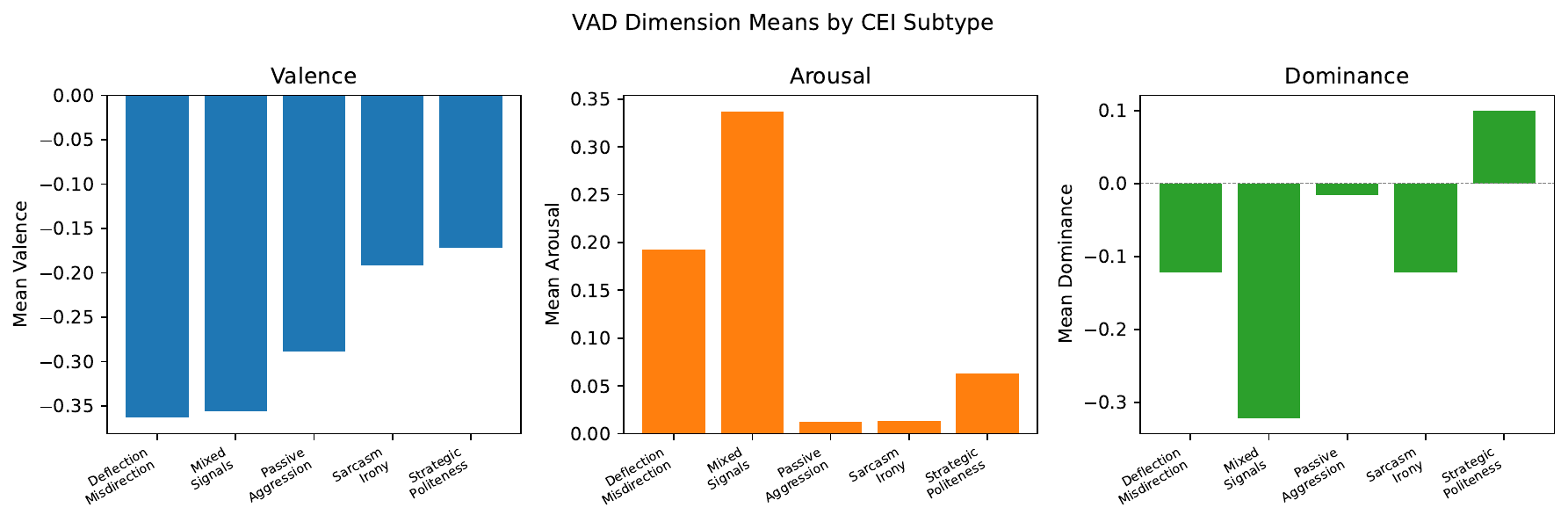}
    \caption{Distribution of VAD ratings across all 900 annotations. Valence skews negative (median $\approx -0.33$), reflecting the dataset's focus on pragmatically complex utterances. Arousal and dominance show broader spread consistent with the diversity of scenarios across power relations and subtypes.}
    \label{fig:vad-distributions}
\end{figure}

The agreement patterns across annotators provide additional context for interpreting these results. Figure~\ref{fig:agreement-subtype} shows the proportion of unanimous, majority, and split scenarios by subtype. Sarcasm/irony shows the highest proportion of unanimous agreement, while deflection produces the most three-way splits, consistent with its near-chance $\kappa$ (0.06). These patterns confirm that the difficulty gradient across subtypes is a robust property of the data, not an artifact of particular annotators. The VAD dimensional data (Figure~\ref{fig:vad-distributions}) further contextualizes the findings: valence ratings skew negative across the dataset (median $\approx -0.33$), reflecting the prevalence of negative or ambiguous affect in pragmatically complex utterances, while arousal and dominance show broader spread consistent with the diversity of social contexts and power configurations.

\begin{figure}[t]
    \centering
    \includegraphics[width=\columnwidth]{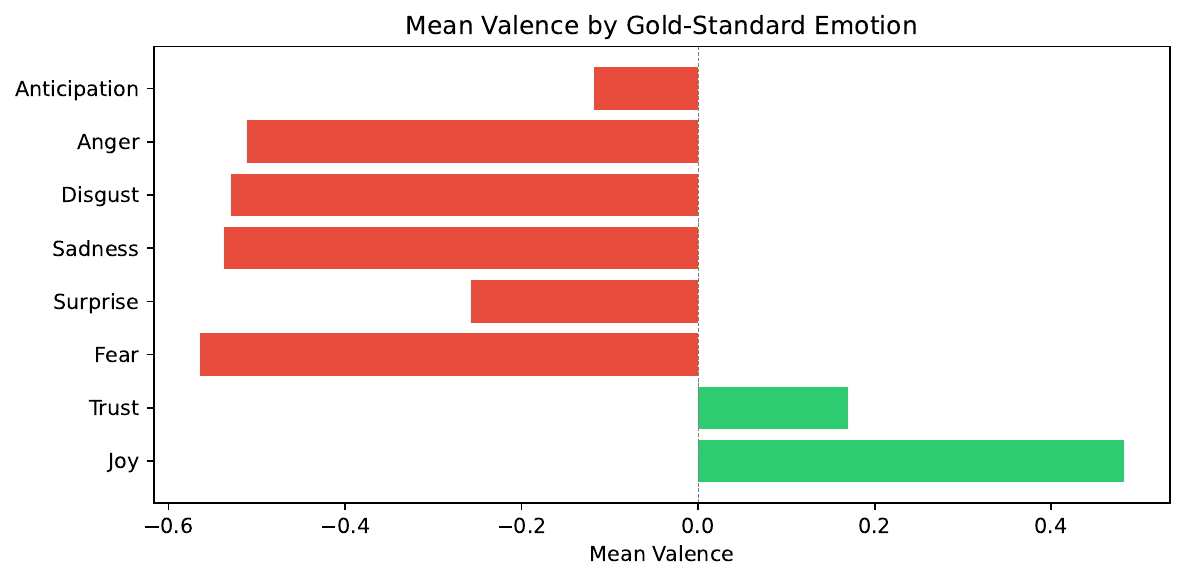}
    \caption{Mean valence by gold-standard emotion category. Positive emotions (joy, trust, anticipation) cluster at positive valence; negative emotions (fear, sadness, anger, disgust) cluster at negative valence; surprise spans both. Error bars show 95\% CIs. The clear separation validates internal consistency of the annotations.}
    \label{fig:emotion-valence}
\end{figure}

The consistency between categorical and dimensional annotations provides a validity check on the data. Figure~\ref{fig:emotion-valence} shows that positive emotions (joy: $\bar{V}=+0.48$; trust: $\bar{V}=+0.17$) receive positive valence ratings, negative emotions (fear: $\bar{V}=-0.56$; sadness: $\bar{V}=-0.54$; anger: $\bar{V}=-0.51$) receive negative valence, and this holds across all subtypes. The clear separation suggests that even when annotators disagree on the emotion category, their underlying affective judgments are coherent---a finding consistent with \citet{pavlick2019inherent}'s observation that disagreement in subjective tasks is often systematic rather than random.

Taken together, these results establish that CEI is a genuinely difficult benchmark that exposes meaningful gaps in current LLM capabilities. The 30-percentage-point gap between models and humans on pragmatic emotion inference is substantially larger than the gaps reported for related tasks such as sentiment analysis or basic emotion classification \citep{demszky2020goemotions}, suggesting that pragmatic reasoning represents a distinct and underserved challenge.

Three observations confirm that CEI has diagnostic utility beyond simply showing that models perform poorly. First, the benchmark \textit{discriminates between models}: accuracy ranges from 13.0\% (Gemini 2.5 Flash) to 25.0\% (Llama-3.1-70B), a 12-percentage-point spread, and models differ most sharply on specific subtypes (e.g., Llama-3.1-70B reaches 33.3\% on mixed signals while Gemini 2.5 Flash scores 13.3\%, but Claude leads on passive aggression). Second, the dissociation between human and model difficulty across subtypes (Spearman $\rho = -0.50$ between human $\kappa$ and mean model accuracy) implies that model errors reflect qualitatively different processing failures rather than simply mirroring human uncertainty. Third, the benchmark occupies a ``diagnostic sweet spot'': the gap between random performance (12.5\%) and the best model (25.0\%) is 12.5 percentage points, while the gap between models and human majority agreement (54\%) is 29 percentage points---leaving substantial room for improvement while confirming that models capture some pragmatic signal.

\begin{figure}[t]
    \centering
    \includegraphics[width=0.85\columnwidth]{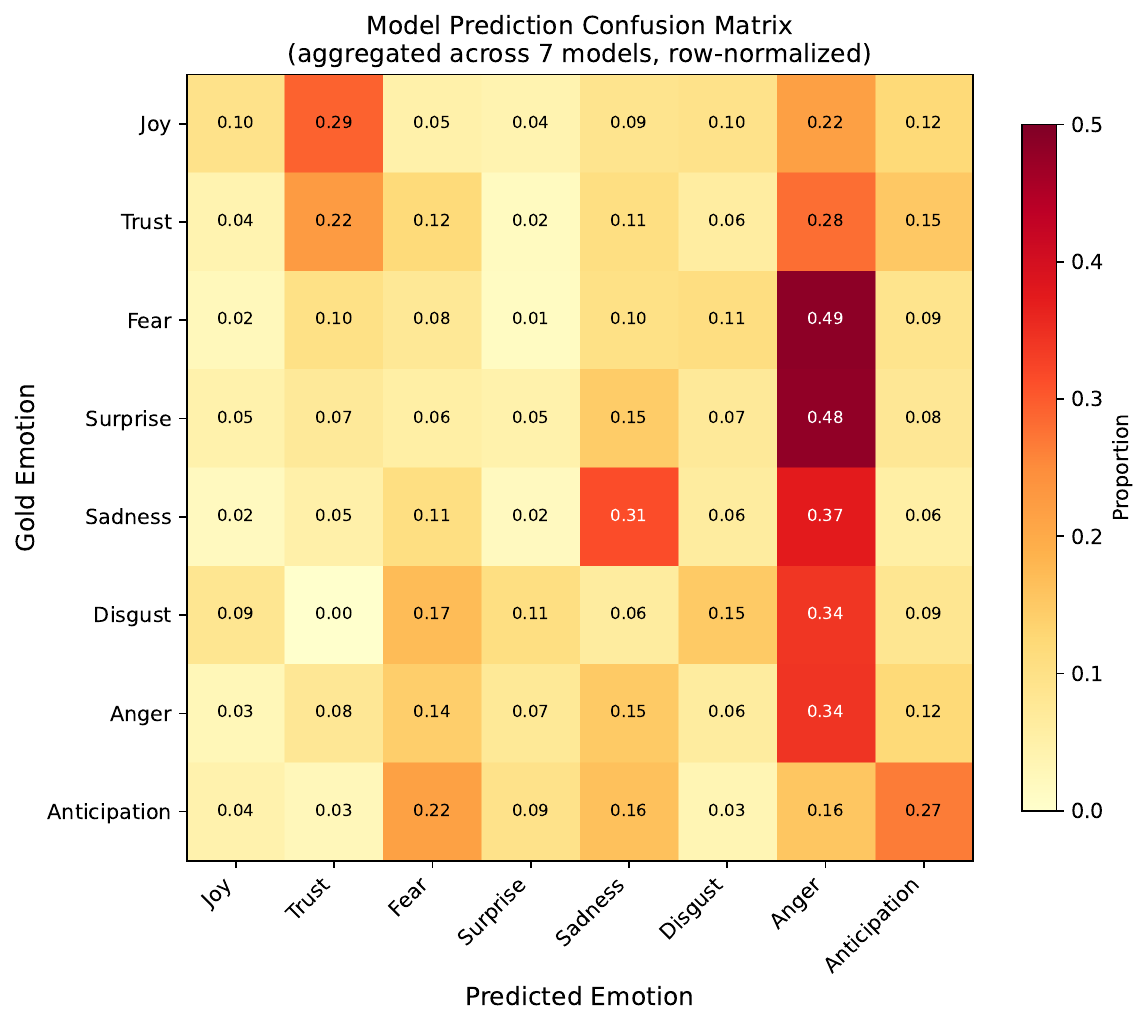}
    \caption{Model prediction confusion matrix aggregated across all 7 baseline models (zero-shot, row-normalized). Models over-predict anger and sadness relative to rarer emotions. The human confusion matrix (Figure~\ref{fig:confusion}) shows a more distributed error pattern.}
    \label{fig:model-confusion}
\end{figure}

The systematic nature of human disagreement and the internal consistency of VAD ratings further indicate that CEI measures something real about pragmatic competence rather than simply adding noise. Figure~\ref{fig:model-confusion} shows the model-level confusion matrix, aggregated across all 7 models in the zero-shot condition. Comparing this with the human confusion matrix (Figure~\ref{fig:confusion}) reveals overlapping but distinct error patterns. Models, like humans, frequently confuse anger and surprise. But models also over-predict anger and sadness at the expense of rarer emotions (disgust, anticipation), concentrating errors more narrowly than humans do. When uncertain, models appear to default to high-frequency negative emotions; human annotators spread their disagreements more evenly across the emotion wheel. The released per-scenario predictions and analysis scripts support further investigation.\footnote{Run \texttt{python scripts/run\_pipeline\_dmlr2026.py --stage all} to regenerate all baseline predictions and confusion matrices.} We additionally note that VAD prediction could serve as a complementary continuous evaluation target, enabling regression-based metrics that capture partial credit for predictions in the right affective neighborhood.

\section{Intended Uses and Limitations}
\label{sec:limitations}

\subsection{Intended Uses}

We designed CEI for three primary uses. First, \textit{benchmarking pragmatic competence}: CEI scores provide a direct measure of whether a model can move beyond surface semantics to infer emotional meaning from indirect speech, a capability that underlies effective social AI. Second, \textit{diagnosing failure modes}: the subtype-level breakdown reveals specific blind spots (e.g., a model that handles mixed signals but fails on sarcasm has a different deficit profile than one that fails uniformly), enabling targeted improvement. Third, \textit{enabling downstream research}: the released data supports several testable hypotheses---do models that score higher on CEI also produce better dialogue responses in social chatbot settings? Can CEI subtype profiles diagnose specific failure modes in deployed systems (e.g., HR screening tools that miss passive aggression, or mental health chatbots that fail to detect distress masked by deflection)? Does fine-tuning on CEI data improve performance on adjacent tasks such as SocialIQa or GoEmotions? The per-annotator labels additionally support soft-label training and calibration-aware evaluation \citep{uma2021learning}.

\subsection{Limitations}

Several design choices constrain the scope and generalizability of CEI. All 300 scenarios are researcher-authored rather than drawn from naturalistic conversation. This is a deliberate methodological choice with both advantages and limitations. The advantages are substantial: expert authoring ensures balanced coverage of all 5 subtypes and 3 power relations, which naturalistic data cannot guarantee; it isolates the target pragmatic phenomenon from confounds present in real conversations (disfluencies, incomplete context, culturally specific idioms); and it has strong precedent in the evaluation literature---the Winograd Schema Challenge \citep{levesque2012winograd} (273 expert-authored items), COPA \citep{roemmele2011choice} (1,000 expert-authored items), and many BigBench tasks \citep{srivastava2022beyond} all use expert-curated stimuli and have proven durable as evaluation instruments. The limitation is that expert-authored scenarios may miss pragmatic patterns that arise only in spontaneous interaction. A future naturalistic extension would complement CEI by testing whether the capability taxonomy established here transfers to in-the-wild data. The dataset covers only English; pragmatic conventions vary substantially across languages and cultures \citep{levinson1983pragmatics}, and we make no claims about cross-linguistic validity. A future multilingual extension would need to revisit both the pragmatic subtypes (which may not carve the space identically in other languages) and the annotation guidelines.

The annotation process introduces its own limitations. All 15 annotators are undergraduate students at a single institution. While this ensures a consistent cultural and linguistic background that supports valid within-dataset comparisons, it limits generalizability; a replication with annotators from diverse backgrounds, including trained NLP researchers or clinical psychologists familiar with pragmatic inference, would strengthen the claim that low agreement reflects task difficulty rather than annotator inexperience. We view this as the most important direction for future validation of CEI. The detailed guidelines (Appendix~\ref{app:guidelines}) consistently ask annotators to label the \textit{speaker's} emotion, but an introductory overview used ``listener'' framing at a high level. Annotators had the detailed guidelines open during annotation, so the speaker framing likely dominated, but we cannot rule out partial contamination from the overview, which may contribute to some disagreement. More broadly, fair agreement ($\kappa = 0.21$) is a property of the task rather than evidence of poor annotation: pragmatic utterances support multiple valid readings, and ground truth labels represent majority consensus, not definitive truth. A model that disagrees with the gold label may sometimes be offering a reasonable alternative interpretation. We encourage researchers to use the released per-annotator labels to evaluate against the full distribution of human judgments, not only the majority vote.

Deflection/misdirection deserves special comment as the hardest subtype ($\kappa = 0.06$, 95\% CI [$-$0.02, 0.14], with only 48\% majority agreement). We view this as a genuine finding rather than a data quality problem: deflection, by definition, involves a speaker who is actively obscuring their emotional state, and annotators facing a deliberately opaque utterance are expected to disagree on which emotion lies behind it. Even so, the meta-annotator's notes reveal that annotators often agreed on valence polarity (e.g., ``all disagree, BUT all have negative valence''), suggesting a shared affective reading beneath the categorical disagreement. We retain all 60 deflection scenarios precisely because they probe the boundary of what discrete emotion labels can capture.

Finally, the dataset's scale and composition reflect deliberate trade-offs. Power relations skew toward peer interactions (72\% peer, 20\% high$\rightarrow$low, 7\% low$\rightarrow$high), mirroring everyday communication but limiting power-stratified analyses. At 300 scenarios (900 annotations), CEI is smaller than crowd-sourced benchmarks like SocialIQa (38K) or SARC (1M), but we chose expert annotation with multi-annotator validation over scale. A chi-squared power analysis ($\alpha=.05$, power$=.80$, $df=7$) gives a detectable Cohen's $w=0.217$ at $N=300$, sufficient to detect distributional departures from chance. What we add over larger datasets is a rigorous validation pipeline and fully documented inter-annotator agreement.

\subsection{Broader Impact}

Better pragmatic inference could help AI systems in mental health support (detecting distress in indirect language), conflict mediation (identifying hostility masked by politeness), and accessibility (interpreting sarcasm and indirect speech for individuals with social communication difficulties). The same capability could be misused for surveillance of employee sentiment, political manipulation through tailored indirect messaging, or deceptive AI agents that exploit power dynamics. We release under CC-BY-4.0 to keep the work open to community scrutiny, and we document all limitations to discourage inappropriate deployment.

\subsection{Reproducibility}

All code, data, and configuration files needed to reproduce the analyses in this paper are released under the MIT license at the project repository.\footnote{\url{https://github.com/jon-chun/cei-tom-dataset-base}} The complete local analysis pipeline (annotation statistics, agreement computation, and figure generation) can be reproduced by running \texttt{python scripts/run\_pipeline\_dmlr2026.py -{}-stage all\_local} ($\sim$30 seconds, no API calls required). LLM baseline results require API access but are fully specified by the prompt templates in Appendix~\ref{app:prompts} and the model configuration in \texttt{config/config-dmlr.yml}. All stochastic operations use seed=42 by default. A detailed reproducibility checklist is provided in Appendix~\ref{app:reproducibility}.

\subsection{Author Responsibility and Data License}

The authors bear all responsibility for the CEI dataset and confirm that its use does not violate any rights. The dataset is released under CC-BY-4.0 and all accompanying code under the MIT license. All scenarios are synthetic (researcher-authored); no real conversations, personal data, or copyrighted materials are included.

\subsection{Maintenance and Versioning}

CEI follows semantic versioning (v1.0.0). The dataset is persistently archived on Zenodo \citep{chun2026cei} and mirrored on the project GitHub repository (see footnote~1). Corrections will be logged in a changelog, and we commit to maintaining the dataset for at least 5 years post-publication.

\section{Conclusion}
\label{sec:conclusion}

We presented CEI, a benchmark of 300 scenarios for evaluating how well language models interpret pragmatically complex utterances in social contexts. The dataset spans five pragmatic subtypes, three power relations, and four social settings, with three independent annotations per scenario. Inter-annotator agreement ($\kappa = 0.06$--$0.25$ by subtype) is low because the task is genuinely hard: reasonable annotators disagree on the emotional meaning of indirect speech, and this disagreement is itself informative. Current LLMs perform poorly (25\% best accuracy vs.\ 54\% human majority agreement), confirming that pragmatic reasoning remains a major gap in language model capabilities. The dissociation between human and model difficulty across subtypes---sarcasm is easiest for humans but hardest for models; deflection shows the reverse---suggests that model errors reflect qualitatively different processing failures rather than simply mirroring human uncertainty.

Several directions for future work follow. The released per-annotator labels enable evaluation against the full distribution of human judgments, supporting soft-label training and calibration-aware metrics \citep{uma2021learning}. The VAD dimensional annotations provide a continuous evaluation target that could complement the categorical task. Multilingual extensions would test whether the pragmatic subtypes and difficulty patterns generalize across cultures. Probing experiments on intermediate model representations could reveal whether LLMs encode pragmatic features internally even when they fail to produce correct predictions. We release CEI under CC-BY-4.0 at \url{https://github.com/jon-chun/cei-tom-dataset-base}.

\impact{
CEI supports research on pragmatic inference in AI systems. Improved pragmatic reasoning could benefit several application domains: mental health chatbots that need to detect distress signals in indirect language (e.g., a patient saying ``I'm fine'' when they are not); accessibility tools for individuals with autism spectrum conditions or social communication difficulties who may struggle to interpret sarcasm, passive aggression, or strategic politeness; conflict mediation platforms that must identify when participants are expressing hostility through superficially polite or compliant language; and educational systems that assess student engagement through indirect cues rather than explicit feedback.

However, the same capabilities pose risks if deployed without appropriate safeguards. Systems that can reliably detect sarcasm, passive aggression, or deflection could be used for workplace surveillance (monitoring employee sentiment from Slack messages), political manipulation (identifying and exploiting emotional vulnerabilities through tailored messaging), or deceptive conversational agents that use indirect speech strategically to influence users. The dataset's focus on power relations adds a further concern: models trained to recognize how power dynamics modulate pragmatic strategy could be used to craft messages that exploit authority asymmetries.

We mitigate these risks through several mechanisms: open release under CC-BY-4.0 (enabling community scrutiny of both the data and models built on it), comprehensive documentation of limitations and failure modes, explicit guidance against high-stakes deployment without human oversight, and the inclusion of all individual annotator labels (rather than just gold standards) to foreground the inherent subjectivity of pragmatic interpretation. We urge researchers building on CEI to consider these dual-use implications and to evaluate potential downstream harms alongside technical performance.
}

\acks{
We thank all co-authors for their contributions to scenario annotation and quality review. Hannah Sussman served as meta-annotator, conducting expert adjudication of all flagged scenarios. Murathan Kocaman, Kirill Sidorko, and Abhigya Koirala made substantial contributions to scenario design, annotation protocol refinement, and quality control beyond their annotation duties. This research received no external funding. The authors declare no competing interests.
}

\bibliography{dmlr2026_cei-tom_dataset}

\clearpage

\appendix
\section{Datasheet for CEI}
\label{app:datasheet}

Following \citet{gebru2021datasheets}, we provide a complete datasheet.

\subsection{Motivation}
\begin{itemize}[leftmargin=*,noitemsep]
    \item \textbf{Purpose:} Evaluate emotion inference from indirect speech in language models.
    \item \textbf{Creators:} Academic research team.
    \item \textbf{Funding:} None; completed as university coursework.
\end{itemize}

\subsection{Composition}
\begin{itemize}[leftmargin=*,noitemsep]
    \item \textbf{Instances:} 300 scenarios with situational context and utterance
    \item \textbf{Labels:} 3 annotations per scenario (emotion + VAD)
    \item \textbf{Splits:} Predefined stratified train/val/test (70/15/15) balanced by subtype and power relation, generated with seed=42. Available in \texttt{reports/dmlr2026/splits.json}
    \item \textbf{Sensitive content:} None; all scenarios are synthetic
\end{itemize}

\subsection{Collection Process}
\begin{itemize}[leftmargin=*,noitemsep]
    \item \textbf{Scenarios:} Template-based synthesis with expert curation; 600 initial scenarios were narrowed to 300 after manual review for quality, clarity, and subtype balance
    \item \textbf{Annotations:} Collected from 15 undergraduate annotators in a university course setting
    \item \textbf{Timeline:} November--December 2025
    \item \textbf{Ethics:} IRB-exempt (Category 2). Annotators received course credit and were offered co-authorship for high-quality contributions
\end{itemize}

\subsection{Preprocessing}
\begin{itemize}[leftmargin=*,noitemsep]
    \item Quality pipeline applied (schema, statistical, agreement, expert adjudication)
    \item 15.7\% of scenarios required expert adjudication
    \item Ground truth established via majority vote + expert override
\end{itemize}

\subsection{Uses}
\begin{itemize}[leftmargin=*,noitemsep]
    \item \textbf{Intended:} Research on pragmatic AI evaluation.
    \item \textbf{Not recommended:} High-stakes decision-making without human oversight.
\end{itemize}

\subsection{Author Responsibility}
\begin{itemize}[leftmargin=*,noitemsep]
    \item The authors bear all responsibility in case of violation of rights and confirm that all data is released with appropriate licensing.
    \item All scenarios are synthetic; no real conversations, personal data, or copyrighted materials are included.
\end{itemize}

\subsection{Distribution}
\begin{itemize}[leftmargin=*,noitemsep]
    \item \textbf{License:} CC-BY-4.0 (data), MIT (code)
    \item \textbf{Code:} \url{https://github.com/jon-chun/cei-tom-dataset-base}
    \item \textbf{Dataset:} \url{https://huggingface.co/datasets/jonc/cei-benchmark}
    \item \textbf{Persistent archive:} Zenodo (DOI: \texttt{10.5281/zenodo.18528706})
    \item \textbf{Machine-readable metadata:} Croissant format (auto-generated by HuggingFace Hub)
\end{itemize}

\subsection{Maintenance}
\begin{itemize}[leftmargin=*,noitemsep]
    \item \textbf{Support:} 5 years minimum
    \item \textbf{Hosting:} GitHub (primary) + Zenodo (persistent archive with DOI)
    \item \textbf{Contact:} \texttt{chunj@kenyon.edu}
\end{itemize}

\section{Annotation Guidelines}
\label{app:guidelines}

This appendix reproduces the annotation protocol provided to annotators. The Label Studio interface used for data collection is shown in Figure~\ref{fig:labelstudio} (\S\ref{sec:annotation}).

\subsection{Core Task Definition}

Annotators are asked: \textit{``Given this context and utterance, what emotion does the SPEAKER most likely feel?''} The target is the speaker's emotional state as inferred from the situation, roles, and utterance.

\subsection{Emotion Categories}

Annotators select from Plutchik's 8 basic emotions:

\begin{itemize}[leftmargin=*,noitemsep]
    \item \textbf{Joy:} Happiness, contentment, amusement
    \item \textbf{Trust:} Acceptance, confidence in the speaker
    \item \textbf{Fear:} Anxiety, worry, apprehension
    \item \textbf{Surprise:} Unexpected information, confusion
    \item \textbf{Sadness:} Disappointment, hurt, dejection
    \item \textbf{Disgust:} Revulsion, contempt, disapproval
    \item \textbf{Anger:} Frustration, irritation, offense
    \item \textbf{Anticipation:} Expectation, interest, engagement
\end{itemize}

\subsection{VAD Ratings}

Annotators also rate each scenario on three 7-point scales mapped to $[-1.0, +1.0]$:

\begin{itemize}[leftmargin=*,noitemsep]
    \item \textbf{Valence} ($-1.0$=very negative, $+1.0$=very positive): How pleasant/unpleasant the speaker feels
    \item \textbf{Arousal} ($-1.0$=very calm, $+1.0$=very activated): The speaker's energy and activation level
    \item \textbf{Dominance} ($-1.0$=very submissive, $+1.0$=very dominant): How in-control the speaker feels
\end{itemize}

\subsection{Subtype Definitions}

\paragraph{Sarcasm/Irony.} The speaker says the opposite of what they mean, often with mocking or critical intent. \textit{Key signal:} Literal interpretation contradicts context.

\paragraph{Mixed Signals.} Verbal content conflicts with contextual cues (tone, situation, body language described). \textit{Key signal:} ``I'm fine'' when clearly not fine.

\paragraph{Strategic Politeness.} Surface-level politeness masks criticism, disagreement, or negative evaluation. \textit{Key signal:} Formally polite language with subtle digs.

\paragraph{Passive Aggression.} Indirect expression of hostility through apparent compliance or withdrawal. \textit{Key signal:} Compliance that feels punishing.

\paragraph{Deflection/Misdirection.} Avoiding uncomfortable topics by changing subject or redirecting attention. \textit{Key signal:} Non-sequitur responses to direct questions.

\subsection{Edge Case Guidance}

\paragraph{Multiple emotions apply.} Select the \textit{primary} or \textit{strongest} emotion. If genuinely equal, choose the more negative emotion (pragmatic ambiguity often signals social friction).

\paragraph{Ambiguous scenarios.} Some scenarios are intentionally ambiguous. Annotate based on your best interpretation; disagreement is expected and informative.

\paragraph{Cultural variation.} Interpret from a general American English perspective. Acknowledge that interpretations may vary across cultures.

\paragraph{Speaker emotion.} Always focus on what the \textit{speaker} most likely feels in the given context, based on the situation, roles, and utterance.

\subsection{Calibration Examples}

These examples were provided in the guidelines to calibrate annotator judgments.

\paragraph{Example 1: Sarcasm vs.\ passive aggression.} \textit{Context:} A team member who was not consulted on a decision says to the group: ``Well, I'm sure you all made the best choice without me.'' \textit{Calibrated label:} \textbf{anger} (passive aggression). The surface compliance might suggest sarcasm, but the indirect hostility through deference points to passive-aggressive anger.

\paragraph{Example 2: Trust vs.\ fear boundary.} \textit{Context:} An employee responds to a new assignment from their supervisor: ``I'll figure it out---I always do.'' \textit{Calibrated label:} \textbf{trust} (self-confidence). The key signal is self-directed confidence (``I always do''), not anxiety about the unknown.

\paragraph{Example 3: Three-way split scenario.} \textit{Context:} At a family dinner, after extended discussion about a sibling's career change, the speaker says: ``That's brave.'' \textit{Calibrated label:} \textbf{no consensus}: trust (genuine admiration), fear (anxiety on the sibling's behalf), and surprise (unexpectedness) are all defensible. This is what a genuine three-way split looks like: the ambiguity is in the utterance, not in the annotation.

\subsection{Quality Expectations}

\begin{itemize}[leftmargin=*,noitemsep]
    \item Median completion time: 30--60 seconds per scenario
    \item Read full context before responding
    \item Use confidence ratings honestly (low confidence is informative)
    \item Flag scenarios that seem malformed or unclear
\end{itemize}

\section{Reproducibility Checklist}
\label{app:reproducibility}

Per the \citet{pineau2021improving} ML Reproducibility Checklist:

\begin{enumerate}[leftmargin=*,noitemsep]
    \item \textbf{Dataset documentation:} Datasheet (Appendix~\ref{app:datasheet}), data dictionary (README), preprocessing steps documented. \checkmark
    \item \textbf{Code availability:} Full pipeline, prompt templates, and figure generation scripts released under MIT license. \checkmark
    \item \textbf{Dependencies specified:} \texttt{pyproject.toml} with pinned versions. \checkmark
    \item \textbf{Installation instructions:} README with setup, run, and verification commands. \checkmark
    \item \textbf{Evaluation metrics defined:} Accuracy, macro-F1, with stratification by subtype and power relation. \checkmark
    \item \textbf{Prompt templates provided:} Zero-shot and chain-of-thought templates in \texttt{prompts/}. \checkmark
    \item \textbf{Ground truth derivation:} Majority vote + expert adjudication, documented in \S\ref{sec:annotation}. \checkmark
    \item \textbf{Confidence intervals:} Bootstrap CIs (2,000 resamples, seed=42) for all $\kappa$ values. \checkmark
    \item \textbf{Inter-annotator agreement:} Fleiss' $\kappa$ per subtype with interpretation (Table~\ref{tab:agreement}). \checkmark
    \item \textbf{Annotator qualifications:} 15 undergraduate annotators, training protocol described (\S\ref{sec:annotation}). \checkmark
    \item \textbf{Quality control:} Validation pipeline with four stages (schema, statistical, agreement, expert). \checkmark
    \item \textbf{Data splits:} Stratified train/val/test (70/15/15) with seed=42. \checkmark
    \item \textbf{Random seeds:} All stochastic operations use seed=42 by default. \checkmark
    \item \textbf{Ethical review:} IRB-exempt (Category 2); annotators received course credit. \checkmark
    \item \textbf{License:} CC-BY-4.0 (data), MIT (code). \checkmark
\end{enumerate}

\noindent\textbf{Replication steps:}

\begin{enumerate}[leftmargin=*,noitemsep]
    \item \textbf{Clone and install:}
\begin{small}
\begin{verbatim}
git clone https://github.com/jon-chun/
    cei-tom-dataset-base.git
cd cei-tom-dataset-base
uv venv && source .venv/bin/activate
uv pip install -e ".[dev,llm]"
\end{verbatim}
\end{small}
    \item \textbf{Local analysis} (no API keys, $\sim$30 seconds):
\begin{small}
\begin{verbatim}
python scripts/run_pipeline_dmlr2026.py \
    --stage all_local
\end{verbatim}
\end{small}
    \item \textbf{LLM baselines} (requires API keys in environment):
\begin{small}
\begin{verbatim}
export OPENAI_API_KEY=... ANTHROPIC_API_KEY=...
export XAI_API_KEY=... GOOGLE_API_KEY=...
export TOGETHER_API_KEY=... FIREWORKS_API_KEY=...

python scripts/run_pipeline_dmlr2026.py \
    --stage run_baselines --prompt-mode zero-shot
python scripts/run_pipeline_dmlr2026.py \
    --stage run_baselines --prompt-mode cot
python scripts/run_pipeline_dmlr2026.py \
    --stage run_baselines --prompt-mode few-shot
\end{verbatim}
\end{small}
    Use \texttt{--resume} to continue from the last completed model if interrupted. Model definitions and pricing are in \texttt{config/config-dmlr.yml}.
    \item \textbf{Generate outputs} (tables, figures, analysis):
\begin{small}
\begin{verbatim}
python scripts/run_pipeline_dmlr2026.py \
    --stage generate_outputs
python scripts/generate_model_confusion_matrix.py
\end{verbatim}
\end{small}
\end{enumerate}

\section{Prompt Templates}
\label{app:prompts}

We provide three prompt templates for CEI evaluation. The \textbf{zero-shot} template presents the scenario and requests a single emotion label. The \textbf{chain-of-thought} template adds explicit reasoning steps before the label. In the \textbf{few-shot} template, three worked examples precede the target scenario.

\subsection{Zero-Shot Prompt (Baseline)}

\begin{small}
\begin{verbatim}
You are evaluating a communication scenario.
Based on the context, determine the primary
emotion the SPEAKER is most likely experiencing
when they make this utterance.

SCENARIO
Situation: {situation}
Speaker ({speaker_role}) says to
  Listener ({listener_role}):
"{utterance}"

Choose exactly one of: joy, trust, fear,
surprise, sadness, disgust, anger, anticipation

Respond with ONLY a JSON object:
{"emotion": "<one of the 8 emotions>"}
\end{verbatim}
\end{small}

\subsection{Chain-of-Thought Prompt}

The CoT prompt uses the same scenario format but replaces the JSON-only instruction with five explicit reasoning steps: (1) literal meaning, (2) contextual cues, (3) pragmatic interpretation, (4) speaker's internal state, and (5) primary emotion selection. The model provides free-text reasoning before answering with ``Answer: [emotion]''.

\subsection{Few-Shot Prompt (3-Shot)}

The few-shot prompt prepends three worked examples (one each of sarcasm/irony, deflection, and strategic politeness) before the target scenario. Each example includes the full scenario text and a gold-standard JSON answer. The target scenario uses the same JSON response format as the zero-shot prompt. All three prompt templates are included in the released repository.

\section{Sample Data}
\label{app:sample-data}

Table~\ref{tab:sample-data} shows three representative annotation records from the released CSVs, one per agreement level (unanimous, majority, split).

\begin{table}[ht]
\centering
\caption{Sample annotation records illustrating the three agreement levels. Each row shows one scenario with the three annotators' emotion labels and the resulting gold standard.}
\label{tab:sample-data}
\small
\begin{tabular}{@{}p{1.3cm}p{4.5cm}p{1.3cm}p{1.3cm}p{1.3cm}p{1.2cm}@{}}
\toprule
\textbf{Subtype} & \textbf{Utterance (excerpt)} & \textbf{Ann.~1} & \textbf{Ann.~2} & \textbf{Ann.~3} & \textbf{Gold} \\
\midrule
Sarcasm & ``Oh sure, I love working weekends.'' & anger & anger & anger & anger \\
\addlinespace
Mixed & ``I'm fine, just tired.'' & sadness & sadness & trust & sadness \\
\addlinespace
Deflection & ``Anyway, who wants coffee?'' & surprise & fear & anger & surprise \\
\bottomrule
\end{tabular}
\end{table}

Each scenario in the released CSVs also includes per-annotator VAD ratings (valence, arousal, dominance on 7-point scales mapped to $[-1.0, +1.0]$), annotator confidence scores, and the meta-annotator's adjudication notes where applicable. The complete data schema is documented in the repository README.

\section{Annotation Logistics}
\label{app:logistics}

\paragraph{Timeline.} Annotation took place over 12 days in November--December 2025. Each annotator completed all 60 scenarios of their assigned subtype in a single session lasting approximately 50--70 minutes (median $\approx$ 1 minute per scenario). The meta-annotator adjudication phase took an additional 8 hours spread over 3 days.

\paragraph{Annotator assignment.} We assigned annotators to subtypes to ensure that each subtype received exactly 3 annotators who rated all 60 scenarios. This within-subtype design enables valid per-subtype $\kappa$ computation with consistent rater sets, at the cost of not being able to assess cross-subtype rater consistency.

\paragraph{Compensation.} Annotation was completed as a graded component of IPHS 391 (Kenyon College, Fall 2025). Annotators received course credit for completing the assignment. Annotators whose work met internal consistency thresholds were offered co-authorship on this paper. No monetary compensation was provided. The study was deemed IRB-exempt (Category 2).

\paragraph{Consent.} All annotators were informed that their annotation labels (but not free-text explanations) would be released as part of a public research dataset and potentially included in a publication. Annotators provided informed consent for the publication of their aggregated and individual emotion and VAD labels. Co-authorship terms were communicated before the annotation period began.

\section{Compute Resources}
\label{app:compute}

All baseline experiments used commercial API endpoints with no local GPU computation. We ran the full 7-model baseline suite in three prompt modes (zero-shot, chain-of-thought, few-shot), totaling 6,300 API calls (7 models $\times$ 300 scenarios $\times$ 3 modes). Wall-clock time was approximately 2.5 hours using parallel execution with per-provider rate limiting. Total API cost was approximately \$15 USD at February 2026 pricing. No fine-tuning or training was performed. The analysis pipeline (\texttt{run\_pipeline\_dmlr2026.py}) runs all local stages in under 30 seconds on a standard laptop. Per-model pricing is listed in \texttt{config/config-dmlr.yml}.

\end{document}